\documentclass[]{memtensor}

\usepackage[toc,page,header]{appendix}
\usepackage[utf8]{inputenc} % allow utf-8 input
\usepackage[T1]{fontenc}    % use 8-bit T1 fonts
\usepackage{hyperref}       % hyperlinks
\usepackage{url}            % simple URL typesetting
\usepackage{booktabs}       % professional-quality tables
\usepackage{amsfonts}       % blackboard math symbols
\usepackage{nicefrac}       % compact symbols for 1/2, etc.
\usepackage{microtype}      % microtypography
\usepackage{amsmath}
\usepackage{etoolbox}
\usepackage{lipsum}
\usepackage{titletoc}
\usepackage[table]{xcolor}  % colors + table color support
\usepackage{tablefootnote}  % footnote in table
\usepackage{threeparttable}
\usepackage{array}
\usepackage{tabularx}
\usepackage{enumitem}
\usepackage{listings}
\ifnum\pdfshellescape=0
    \usepackage[frozencache]{minted}
\else
    \usepackage{minted}
\fi
\newif\ifsvFinalizeMarkdownCache
\ifdefined\SVFinalizeMarkdownCache
    \svFinalizeMarkdownCachetrue
\fi
\ifsvFinalizeMarkdownCache
    \usepackage[
        cacheDir=assets/case_study/markdown-cache,
        frozenCacheFileName=assets/case_study/markdown-cache/frozenCache.tex,
        finalizecache
    ]{markdown}
\else
    \usepackage[
        cacheDir=assets/case_study/markdown-cache,
        frozenCacheFileName=assets/case_study/markdown-cache/frozenCache.tex,
        frozencache
    ]{markdown}
\fi
\usepackage{seqsplit}
\usepackage{pmboxdraw}
\usepackage{arydshln}
\usepackage{multirow}
\usepackage{xspace}
\tcbuselibrary{minted,breakable,skins}

\makeatletter
\let\textbf\@undefined
\makeatother
\DeclareTextFontCommand{\textbf}{\fontseries{b}\selectfont}

\IfFileExists{fontawesome5.sty}{
    \usepackage{fontawesome5}
}{
    \newcommand{\faGlobe}{[Web]}
    \newcommand{\faGithub}{[GH]}
    \newcommand{\faEnvelope}{[Email]}
    \newcommand{\faCalendar}{[Date]}
    
}

\newcommand{\ours}{\textsc{SkillsVote}\xspace}
\colorlet{svGainColor}{green!50!black}
\colorlet{svLossColor}{red!70!black}
\colorlet{svPromptBlue}{memblue}
\colorlet{svPromptBlueLight}{memblue2}
\colorlet{svPromptAccentColor}{svPromptBlue!85!black}
\newcommand{\svgain}[1]{\textcolor{svGainColor}{#1}}
\newcommand{\svloss}[1]{\textcolor{svLossColor}{#1}}
\newcommand{\svup}[1]{$_{\svgain{\scriptstyle\uparrow #1}}$}
\newcommand{\svdown}[1]{$_{\svloss{\scriptstyle\downarrow #1}}$}

\title{\ours: Lifecycle Governance of Agent Skills from Collection, Recommendation to Evolution}

\author[1,2*]{Hongyi Liu}
\author[1,2*]{Haoyan Yang}
\author[1,3*]{Tao Jiang}
\author[1]{Bo Tang}
\author[1]{Feiyu Xiong}
\author[4\dagger]{Yuyu Luo}
\author[1\dagger]{Zhiyu Li}

\affiliation[1]{MemTensor (Shanghai) Technology}
\affiliation[2]{Harbin Institute of Technology}
\affiliation[3]{Soochow University}
\affiliation[4]{The Hong Kong University of Science and Technology (Guangzhou)}

\contribution[*]{Co-first authors}
\contribution[\dagger]{Corresponding author}

\abstract{
Long-horizon LLM agents generate traces that could become reusable experience, but raw trajectories are noisy, local, and hard to govern. Agent Skills offer a structured artifact for combining procedural guidance, executable resources, and applicability boundaries. Yet open skill ecosystems contain redundant, uneven, environment-sensitive artifacts, and indiscriminate updates can pollute future context. We present \ours, a lifecycle-governance framework for Agent Skills across collection, recommendation, attribution, and evolution. \ours profiles a million-scale open source corpus for environment requirements, quality, and verifiability, and synthesizes tasks for verifiable skills. Before execution, it performs agentic library search over structured skill folders to expose instructional context. After execution, it decomposes trajectories into skill-linked subtasks, attributes outcomes to skill-guided execution, agent exploration, environment, and result signals, and admits only successful reusable discoveries to evidence-gated updates. Experiments on Terminal-Bench 2.0 and SWE-Bench Pro show that \ours improves agent performance on challenging agentic coding benchmarks. The gains arise from two complementary pathways: online evolution over task streams at test time and offline transfer via frozen libraries built from either historical trajectories or curated open source skills.

}
\checkdata[\faCalendar\ Date]{\today}
\checkdata[\faEnvelope\ Correspondence]{Yuyu Luo (\email{yuyuluo@hkust-gz.edu.cn}), Zhiyu Li (\email{lizy@memtensor.cn})}
\checkdata[\faGlobe\ Website]{\href{https://skills.vote}{skills.vote}}
\checkdata[\faGithub\ GitHub]{\href{https://github.com/MemTensor/skills-vote}{MemTensor/skills-vote}}

\begin{document}
\maketitle
\clearpage
\section{Introduction}
Recent progress in LLM agents has shifted the research focus from single-turn answer generation to systems that act over long horizons. Contemporary benchmarks require agents to repair realistic codebases \citep{jimenez2024swe,deng2025swe}, navigate web applications \citep{zhou2024webarena}, operate across desktop environments \citep{xie2024osworld}, and manipulate external state through APIs \citep{trivedi2024appworld}, tools, and terminals \citep{merrill2026terminal}. These settings produce trajectories of intermediate decisions, tool interactions, and environmental feedback. Prior work on experiential agents shows that such traces can shape later behavior, but only after low-level execution evidence is distilled into reusable experience or skills \citep{shinn2023reflexion,zhao2024expel,wang2024voyager}.

\begin{figure*}[b]
    \centering
    \includegraphics[width=\textwidth]{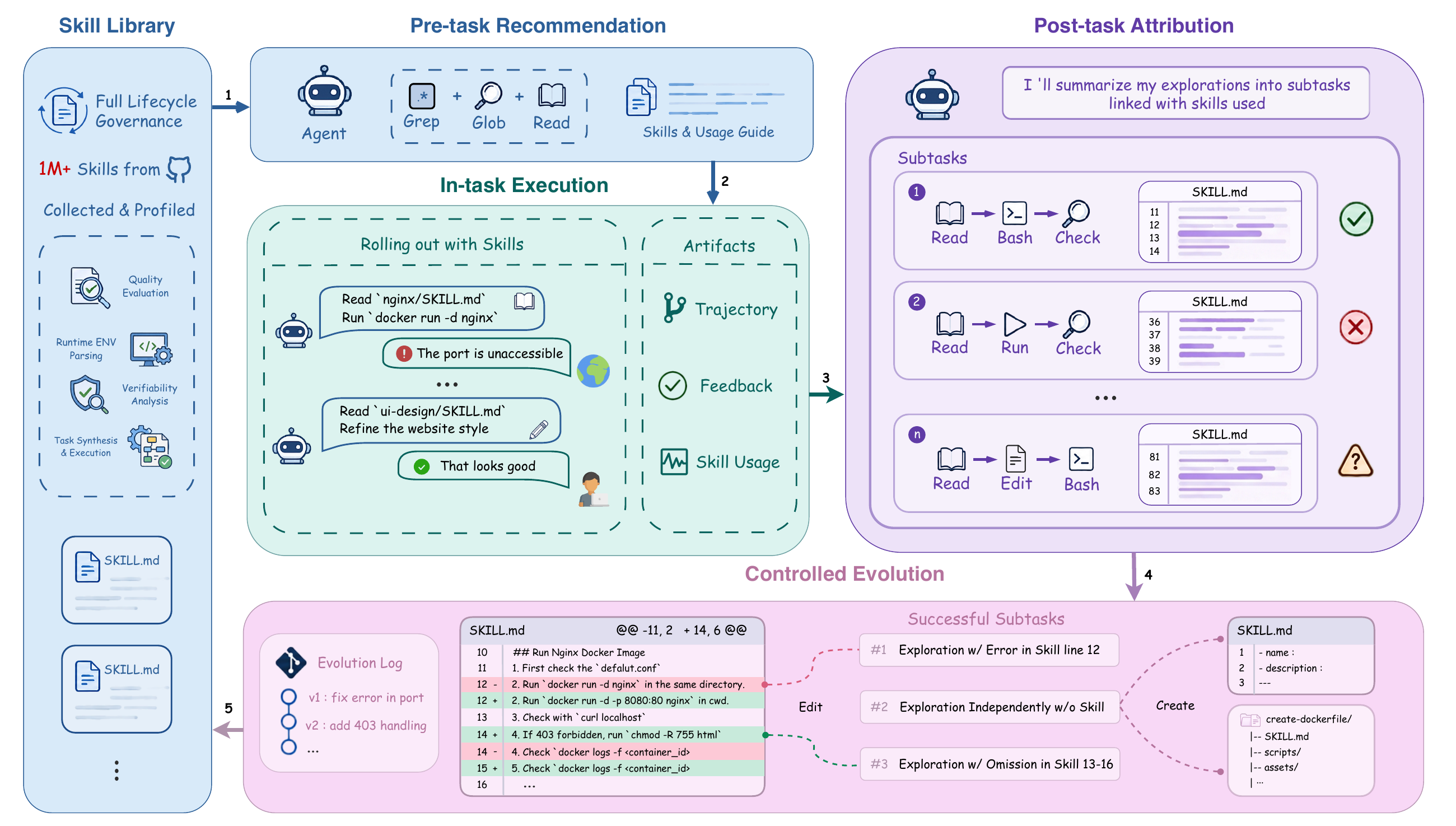}
    \caption{\ours couples pre-task recommendation with post-task attribution and controlled library evolution. A profiled skill library is searched before execution; after execution, trajectories and outcome signals are decomposed into skill-linked subtasks so reusable discoveries can edit existing skills or create new ones.}
    \label{fig:overview}
\end{figure*}

Raw trajectories, however, are a poor substrate for durable experience reuse. They are lengthy, noisy, tightly bound to local environments, and often conflate robust strategies with incidental state. Agent Skills offer a more structured schema: they package procedural instructions, scripts, templates, references, dependency boundaries, and applicability conditions into auditable artifacts, making experience more compact than full trajectories while preserving more executable context than isolated natural language summaries \citep{jiang2026sok}.

At ecosystem scale, the problem is no longer only how to author an individual skill, but how to control a continuously expanding library. Public skill ecosystems already exhibit scale, redundancy, uneven quality, and safety risks \citep{ling2026agent}. Skill benchmarks further show that the benefit of skills depends on task, domain, and retrieval setting; weakly related or poorly written skills can degrade agent performance \citep{li2026skillsbench,liu2026well}. Treating skills as ecosystem artifacts also changes the failure mode: larger libraries increase coverage, but they also enlarge the search space and amplify library pollution when weakly supported lessons are incorporated indiscriminately. These observations suggest that skill ecosystems at scale require collection, governance, profiling, recommendation, evaluation, and evolution to be treated as coupled processes \citep{li2026organizing,zheng2026skillrouter}. Against this background, \ours constructs and profiles a corpus of more than one million open source Agent Skills and governs how skills \textcolor{memblue}{\textit{vote}} into the agent context before execution and how attributed evidence \textcolor{memblue}{\textit{votes}} into the skill library after execution.

We introduce \ours, a lifecycle framework for Agent Skills. Before execution, \ours formulates recommendation as agentic search over a structured skill library rather than static semantic matching \citep{li2026beyond}; it selects a small, relevant skill set with little redundancy and supplies compressed usage context. After execution, \ours performs outcome attribution from the trajectory and visible result signal, explaining how the outcome relates to the selected skill, the agent's own exploration, environmental conditions, and the evaluation signal itself. These stages form the closed loop shown in Figure~\ref{fig:overview}.

Recent systems for skill evolution show that execution evidence can improve skills by distilling lessons from individual trajectories into transferable skill directories \citep{ni2026trace2skill}, aggregating interaction trajectories across users into shared skill updates \citep{ma2026skillclaw}, and diagnosing bad cases to refine domain skills \citep{liu2026skillforge}. \ours uses this evidence under an attribution control layer: it first determines whether a success or failure is attributable to the selected skill, the agent's own exploration, the environment, or the result signal, and then constrains which experience may enter the evolving skill library. This control prevents spurious successes from being rewarded and keeps failures caused by the environment or evaluation signal from driving irrelevant repository edits. Thus, \ours connects recommendation, attribution, and controlled evolution into an auditable closed loop.

We evaluate \ours on Terminal-Bench 2.0 \citep{merrill2026terminal} and SWE-Bench Pro \citep{deng2025swe}. Our experiments study whether recommendation outperforms directly exposing the initial skill library, whether offline evolution transfers from historical Terminal-Bench Pro trajectories to Terminal-Bench 2.0, and whether online evolution accumulates useful experience in a task stream at test time.

This paper makes the following contributions:
\begin{enumerate}
    \item We formulate an Agent Skill lifecycle framework that connects collection and governance for open skill ecosystems, recommendation, outcome attribution, and controlled evolution.
    \item We construct and profile a corpus of more than one million open source Agent Skills for systematic analysis and governance of open skill ecosystems.
    \item We design a loop that uses attribution to connect recommendation and evolution, constraining skill library evolution and reducing the risk of indiscriminate library updates.
    \item We show that \ours improves online evolution, offline transfer, and skill use controlled by recommendation on Terminal-Bench 2.0 and SWE-Bench Pro public.
\end{enumerate}

% \section{Preliminary}
% \input{sections/2_preliminary.tex}

\section{Related Work}
\paragraph{Evolution of Agent Experience Learning.}
Agent experience learning has progressed from records reusable only in context to executable artifacts. Early memory methods store unstructured cases and examples, such as few-shot trajectories, exemplars, or human-curated interaction records \citep{zhou2025memento,zheng2024synapse,wang2026memgovern}. Workflow methods abstract traces into semi-structured workflows and SOPs \citep{wang2025agent,fang2025memp}, while strategy-level methods compress experience into principles, heuristics, and strategies \citep{zhao2024expel,ouyang2026reasoningbank,cao2025remember,zhang2026agentic,cai2025training,cai2025flex,wang2026procedural}. Recent tool, MCP, and methods for skill learning attach experience to callable interfaces, dependencies, and execution boundaries \citep{lu2026beyond,liu2025unifying,huang2025cascade}. Surveys and systems similarly frame memories, rules, skills, protocols, and harness components as deployment-time external artifacts \citep{zhang2026experience,zhou2026externalization,lin2026position,zhang2026autogenesis,liang2026genericagent,lin2026agentic}. \ours focuses on skill libraries: a skill combines procedural text, scripts, dependencies, and applicability boundaries, so experience remains auditable, versionable, and portable across harnesses, while full harness or protocol evolution has a larger action space.

\paragraph{Agent Skill Ecosystems, Retrieval and Evaluation.}
As Agent Skills become installable and shareable file artifacts \citep{agentSkills,anthropic2026claudeCodeSkills,openai2026ChatgptSkillsDocs,skillsmp2026,skillssh2026,openclaw2026SkillsDocs,hermes2026AgentSkillsGuide,openclaw2026clawhub}, the problem shifts from authoring skills to governing and using open ecosystems. AgentSkillOS \citep{li2026organizing} and SkillNet \citep{liang2026skillnet} organize skills as ecosystem objects, while SkillsBench \citep{li2026skillsbench}, SkillCraft \citep{chen2026skillcraft}, SkVM \citep{chen2026skvm}, and SkCC \citep{ouyang2026skcc} show that utility, compositional use, portability, security, dependencies, and harness compatibility must be evaluated before \texttt{SKILL.md} files can be trusted. \ours profiles open-source skills for format, dependency, quality, and verifiability. At task time, governance does not remove selection: providing skills does not ensure correct selection, composition, or use. SkillRouter learns routing over full skill bodies rather than only names or descriptions \citep{zheng2026skillrouter}, while DCI replaces embedding retrieval with direct corpus interaction over source documents \citep{li2026beyond}. \ours lightly applies filesystem-native inspection to governed skill folders and outputs compact guidance for combining the selected skills.

\paragraph{Skill-Centric Agent Self Evolution.}
A growing body of work studies how agents learn and evolve around skill libraries. One line trains policies to decide when to retrieve a skill, how to use it, and when to distill behavior into the model or revise the library \citep{xia2026skillrl,wang2025reinforcement,xia2026metaclaw,wang2026openclaw,lu2026skill0,shi2026skill1,ouyang2026skillos}. Other systems keep the base model fixed and turn coarse session- or trajectory-level evidence, together with verifier or environment feedback, into reusable skill artifacts \citep{ni2026trace2skill,alzubi2026evoskill,zhang2026coevoskills,ma2026skillclaw,wang2026skillx,yang2026autoskill,si2026context,zhang2026skillflow,zhou2026memento,skillpro2026,gong2026skillmoo,xu2026multi}. \ours further factorizes each trajectory into judged, skill-linked subtasks, localizes the skill knowledge actually used and the responsibility for each outcome, and admits only reusable successful exploration into skill library evolution.

\section{Approach}
\ours treats Agent Skills as lifecycle artifacts: given a skill library, it controls which skills enter the solver agent context before execution and which execution evidence is allowed to update the library afterward. We describe corpus collection, profiling, and task synthesis from skills as preprocessing details in Appendix~\ref{app:open-source-skill-corpus}.

\subsection{Skill Recommendation via Agentic Library Search}

Existing skill harnesses commonly rely on progressive disclosure: the solver agent first sees lightweight skill metadata, and the full \texttt{SKILL.md} and supporting resources are loaded only after the skill appears relevant \citep{openai2026codexSkills,anthropic2026claudeCodeSkills,agentSkills}. This design lets many skills coexist in one environment, but it also compresses pre-task selection into short descriptions and limited path cues. SkillRouter further shows that, in large skill pools, the full skill body often carries decisive routing signals \citep{zheng2026skillrouter}. \ours builds on this interface by adding an exposure control layer conditioned on the task before the solver agent starts execution.

The motivation extends beyond skill systems. Filesystem-native atomic tools are increasingly used as a general interface for agentic search: DCI lets agents interact directly with corpora in retrieval for deep research rather than consuming a fixed top-$k$ interface \citep{li2026beyond}; systems for code search, such as CodeScout \citep{sutawika2026codescout} and SWE-grep \citep{pan2025swegrep}, train multi-turn localization over ordinary repository tools; Vercel's studies of data agents compress query exploration and validation into a small set of file and shell operations \citep{qu2025we,goyal2026testing}; and Letta uses a filesystem backend for agent memory retrieval \citep{letta2025benchmarking}. These examples motivate treating an evolving skill library as a searchable file-based substrate.

Given a task and a skill library, \ours runs a separate recommendation stage. The agent does not solve the task. It searches the local skill library, selectively reads candidate \texttt{SKILL.md} files and related resources, and selects skills that cover the task, fit the target environment, and provide complementary guidance. The output is a compact set of exposed skills plus a short usage guide for the solver agent, rather than the full library, a metadata-only routing decision, or a single-step top-$k$ chunk list. The recommendation record also anchors later attribution: after execution, \ours can inspect whether exposed skills were actually used and whether they contributed reusable discoveries.

\begin{figure*}[!h]
    \centering
    \includegraphics[width=\textwidth]{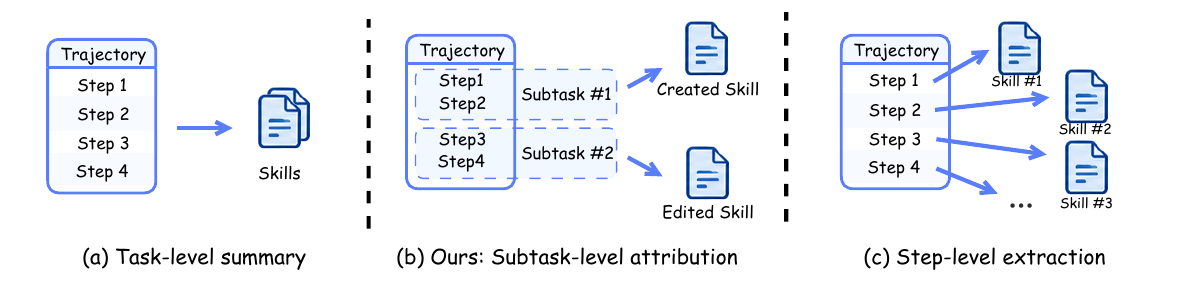}
    \caption{Subtask-level attribution bridges coarse task summaries and fragmented step traces, linking coherent execution segments to skill updates.}
    \label{fig:attribution-granularity}
\end{figure*}

\subsection{Distilling Execution Traces into Evolvable Units}
Recent work exposes a granularity gap in learning from agent execution. Agent evaluation commonly relies on task-level success signals, which are authoritative but provide sparse supervision for long-horizon tool use and make credit assignment difficult \citep{fan2026agentprocessbench}. Meanwhile, systems for skill learning show that execution trajectories contain reusable experience that can improve future behavior \citep{ni2026trace2skill,fang2026trajectory}. However, these works also suggest that trajectories must be filtered before they become reusable artifacts: a run may mix skill-guided actions, independent exploration, corrected failures, and redundant operational steps. At the other end, process-level benchmarks \citep{fan2026agentprocessbench} and failure-diagnosis methods \citep{barke2026agentrx} annotate individual agent steps, showing the value of local feedback for analysis. However, a single tool call rarely constitutes reusable skill knowledge. As shown in Figure~\ref{fig:attribution-granularity}, skill evolution thus requires an intermediate unit between full trajectories and individual steps.

\begin{figure*}[!t]
    \centering
    \includegraphics[width=\textwidth]{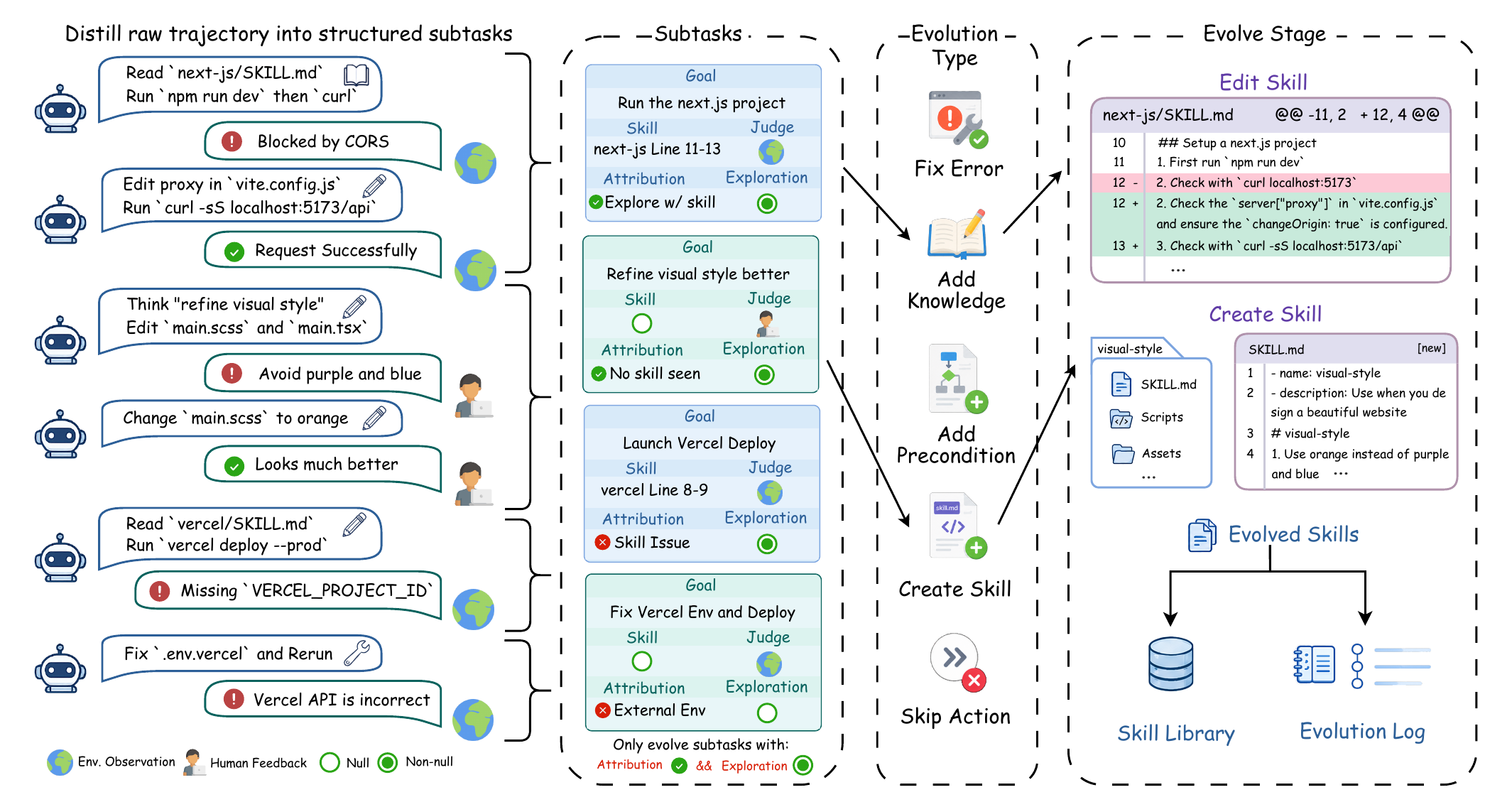}
    \caption{\ours converts execution traces into attributed subtasks, then updates the library only from successful, reusable evidence.}
    \label{fig:attribution-pipeline}
\end{figure*}

\ours addresses this mismatch by inserting a subtask-level attribution layer between full trajectories and individual tool calls. \textbf{A subtask is the smallest semantically complete unit that can support library evolution}: it has \textit{one standalone objective}, \textit{one primary evaluation signal}, and \textit{at most one associated skill context}. The primary evaluation signal specifies what kind of evidence can support the subtask outcome, such as environment feedback, human review, or no explicit signal. Trajectories are split only when one of these three boundaries changes, rather than whenever the agent issues another command. This granularity is local enough to assign responsibility, yet abstract enough to capture reusable procedures, constraints, and recovery patterns.

For each subtask, attribution compresses the execution evidence along three axes:

\begin{enumerate}
    \item \textbf{Outcome evidence.} The system records whether the subtask can be assessed by objective environment feedback, depends on human preference, or lacks an explicit evaluation signal. This prevents verifier-backed outcomes, subjective goals, and unsupported claims from being treated alike.
    \item \textbf{Responsibility assignment.} The system assigns both the final state and its main cause. Successful subtasks may be credited to skill-guided execution, independent exploration, or exploration after observing an irrelevant skill. Failed or uncertain subtasks are retained as diagnostic evidence, but they do not directly authorize skill evolution.
    \item \textbf{Reusable delta.} For skill-related subtasks, the system localizes the portions of skill knowledge that actually shaped execution, rather than crediting every exposed skill. It also extracts only reusable discoveries, such as missing procedures, preconditions, or recovery patterns, while discarding ordinary trial-and-error, task-specific constants, and repetitive operational details.
\end{enumerate}

Together, these fields define an evolvable unit for the trajectory evidence in Figure~\ref{fig:attribution-pipeline}: evidence-bound, responsibility-aware, and reusable. These units form the interface to controlled evolution, where only successful subtasks with reusable exploration can propose library updates.

\subsection{Evidence-Based Controlled Skill Evolution}

The attribution layer produces evolvable units, but library evolution still requires explicit control over what evidence is allowed to change persistent skills. \ours formulates this step as evidence-gated update construction with explicit admissibility, aggregation, and routing criteria.

\paragraph{Admissibility.} \ours first filters which units may trigger evolution. A unit is admissible only if it is successful and contains reusable exploration. Failed, uncertain, or weakly supported evidence may remain useful for diagnosis, but it cannot directly authorize a skill update.
\paragraph{Aggregation.} Admissible units are then grouped before any edit is made. Units that support the same reusable procedure, precondition, workaround, or correction are merged into a single proposed update, so repeated observations strengthen one change rather than producing duplicate or fragmented edits.
\paragraph{Routing.} Finally, \ours routes each aggregated evidence group to an update action. If the evidence extends a skill that actually shaped execution, the system edits that skill through the smallest justified change: fixing incorrect guidance, adding missing knowledge, or tightening prerequisites. If the evidence reflects an independent reusable capability outside the current skill boundary, the system creates a new skill. When evidence is weak, redundant, or semantically misaligned with the target skill, it skips evolution.

Thus, skill evolution is conservative by design: every library change must be supported by attributed execution evidence, localized to the relevant skill boundary, and expressed as reusable procedural knowledge rather than a trajectory recap.

\section{Experiments}
\subsection{Experimental Setup}
We organize the evaluation around three lifecycle questions:
\begin{enumerate}
    \item \textbf{Offline skill transfer.} Can frozen libraries from historical trajectories or curated open source skills improve unseen tasks?
    \item \textbf{Online skill evolution.} Can \ours accumulate useful skills over a sequential task stream?
    \item \textbf{Pre-task recommendation.} Given a growing skill library, does pre-task recommendation reduce negative transfer?
\end{enumerate}

\paragraph{Benchmarks.}
We evaluate with Harbor \citep{harborFramework} on Terminal-Bench 2.0 \citep{merrill2026terminal} and SWE-Bench Pro public \citep{deng2025swe}, two challenging agentic coding benchmarks. We report avg@5 Accuracy on Terminal-Bench 2.0 and avg@1 Resolve Rate on SWE-Bench Pro public, following their leaderboard protocols.

\paragraph{Configurations.}
We run Codex with three model and effort pairs: GPT-5.2 medium \citep{openai2025gpt52}, GPT-5.4 mini medium effort \citep{openai2026gpt54MiniNano}, and GPT-5.5 xhigh \citep{openai2026gpt55}. For each benchmark and backbone pair, we compare three classes of settings: (1) the w/o skills setting is the base solver without an external skill library; (2) Online settings start from an empty experience library and update it along the task stream at test time; we compare \ours with ReasoningBank \citep{ouyang2026reasoningbank} and skill-creator, a direct baseline that converts completed trajectories into reusable skills without edit/create decisions guided by attribution; (3) Offline settings start from a frozen skill library and use it only through pre-task recommendation on the test set. TB-Pro is evolved from historical Terminal-Bench Pro \citep{wang2025let} task trajectories and is used only for Terminal-Bench 2.0 transfer, while Curated is a frozen library of 10K curated open source skills, used only by the recommendation stage. Appendix~\ref{app:experiment-details} gives the complete configuration details.

\begin{table*}[!t]
    \caption{Main results on Terminal-Bench 2.0. Scores are avg@5 Accuracy; deltas denote absolute changes in percentage points from the corresponding baseline without skills.}
    \label{tab:main-results-tb2}
    \centering
    \scriptsize
    \setlength{\tabcolsep}{4pt}
    \renewcommand{\arraystretch}{1.35}
    \resizebox{\textwidth}{!}{%
    \begin{tabular}{
        >{\raggedright\arraybackslash}m{0.24\linewidth}
        >{\raggedright\arraybackslash}m{0.20\linewidth}
        >{\raggedright\arraybackslash}m{0.145\linewidth}
        >{\raggedright\arraybackslash}m{0.17\linewidth}
        >{\raggedright\arraybackslash}m{0.135\linewidth}
    }
        \toprule
        \multirow{2}{*}{\textbf{Settings}} & \multicolumn{1}{c}{\underline{\textbf{Overall}}} & \multicolumn{1}{c}{\textbf{Easy}} & \multicolumn{1}{c}{\textbf{Medium}} & \multicolumn{1}{c}{\textbf{Hard}} \\
        & \multicolumn{1}{c}{(89)} & \multicolumn{1}{c}{(4)} & \multicolumn{1}{c}{(55)} & \multicolumn{1}{c}{(30)} \\
        \specialrule{\lightrulewidth}{\aboverulesep}{0pt}
        \rowcolor{black!8}
        \multicolumn{5}{c}{GPT-5.2 medium} \\
        w/o skills & 51.1 & \underline{75.0} & 54.9 & \underline{40.7} \\
        \multicolumn{5}{l}{Online} \\
        \quad \ours & 53.7\svup{2.6} & \underline{75.0} & \underline{62.9}\svup{8.0} & 34.0\svdown{6.7} \\
        \quad ReasoningBank & 52.1\svup{1.0} & 70.0\svdown{5.0} & 58.6\svup{3.7} & 38.0\svdown{2.7} \\
        \quad skill-creator & 53.7\svup{2.6} & 60.0\svdown{15.0} & 61.5\svup{6.6} & 38.7\svdown{2.0} \\
        \multicolumn{5}{l}{Offline} \\
        \quad TB-Pro & \textbf{58.9}\svup{7.8} & \textbf{90.0}\svup{15.0} & \textbf{65.1}\svup{10.2} & \textbf{43.3}\svup{2.6} \\
        \quad Curated & \underline{54.8}\svup{3.7} & \underline{75.0} & 61.8\svup{6.9} & 39.3\svdown{1.4} \\
        \specialrule{\lightrulewidth}{\aboverulesep}{0pt}
        \rowcolor{black!8}
        \multicolumn{5}{c}{GPT-5.4 mini medium} \\
        w/o skills & 51.7 & \textbf{75.0} & 61.8 & 30.0 \\
        \multicolumn{5}{l}{Online} \\
        \quad \ours & 52.8\svup{1.1} & \textbf{75.0} & 63.6\svup{1.8} & 30.0 \\
        \quad ReasoningBank & 50.6\svdown{1.1} & \underline{70.0}\svdown{5.0} & 62.6\svup{0.8} & 26.0\svdown{4.0} \\
        \quad skill-creator & 47.6\svdown{4.1} & 65.0\svdown{10.0} & 59.6\svdown{2.2} & 23.3\svdown{6.7} \\
        \multicolumn{5}{l}{Offline} \\
        \quad TB-Pro & \textbf{57.5}\svup{5.8} & \underline{65.0}\svdown{10.0} & \underline{64.7}\svup{2.9} & \textbf{43.3}\svup{13.3} \\
        \quad Curated & \underline{55.7}\svup{4.0} & \textbf{75.0} & \textbf{65.5}\svup{3.7} & \underline{35.3}\svup{5.3} \\
        \specialrule{\lightrulewidth}{\aboverulesep}{0pt}
        \rowcolor{black!8}
        \multicolumn{5}{c}{GPT-5.5 xhigh} \\
        w/o skills & 79.8 & 90.0 & 83.0 & \textbf{72.1} \\
        \multicolumn{5}{l}{Online} \\
        \quad \ours & \underline{80.7}\svup{0.9} & \textbf{100.0}\svup{10.0} & \underline{84.9}\svup{1.9} & 70.0\svdown{2.1} \\
        \quad ReasoningBank & 77.7\svdown{2.1} & 85.0\svdown{5.0} & 81.9\svdown{1.1} & 68.6\svdown{3.5} \\
        \quad skill-creator & 78.8\svdown{1.0} & 90.0 & 81.9\svdown{1.1} & \underline{71.4}\svdown{0.7} \\
        \multicolumn{5}{l}{Offline} \\
        \quad TB-Pro & \textbf{81.2}\svup{1.4} & \underline{95.0}\svup{5.0} & \underline{84.9}\svup{1.9} & \textbf{72.1} \\
        \quad Curated & 78.1\svdown{1.7} & \underline{95.0}\svup{5.0} & \textbf{85.3}\svup{2.3} & 62.1\svdown{10.0} \\
        % \midrule
        % \rowcolor{black!8}
        % \multicolumn{5}{c}{Claude Code} \\
        % Claude 4.5 & -- & -- & -- & -- \\
        % \quad \online & -- & -- & -- & -- \\
        % \quad \offline & -- & -- & -- & -- \\
        \bottomrule
    \end{tabular}%
    }
\end{table*}

\begin{table*}[t]
    \caption{Main results on SWE-Bench Pro public. Scores are avg@1 Resolve Rate; deltas denote absolute changes in percentage points in the overall column.}
    \label{tab:main-results-swebenchpro}
    \centering
    \scriptsize
    \setlength{\tabcolsep}{2.5pt}
    \renewcommand{\arraystretch}{1.35}
    \resizebox{\textwidth}{!}{%
    \begin{tabular}{
        >{\raggedright\arraybackslash}m{0.90in}
        w{l}{0.46in}
        *{11}{w{c}{0.34in}}
    }
        \toprule
        \multirow{2}{*}{\textbf{Settings}} & \multicolumn{1}{c}{\underline{\textbf{Overall}}} & \textbf{ansible} & \textbf{openlib.} & \textbf{qutebro.}
        & \textbf{flipt} & \textbf{telepor.} & \textbf{vuls} & \textbf{navidro.}
        & \textbf{webclie.} & \textbf{element.} & \textbf{nodebb} & \textbf{tutanot.} \\
        & \multicolumn{1}{c}{(731)} & (96) & (91) & (79)
        & (85) & (76) & (62) & (57)
        & (65) & (56) & (44) & (20) \\
        \specialrule{\lightrulewidth}{\aboverulesep}{0pt}
        \rowcolor{black!8}
        \multicolumn{13}{c}{GPT-5.2 medium} \\
        w/o skills
        & 47.6 & 49.0 & 64.8 & \underline{62.0} & \textbf{32.9} & \underline{34.2} & \underline{54.8} & \underline{49.1} & \textbf{43.1} & 50.0 & 47.7 & 0.0 \\
        \multicolumn{13}{l}{Online} \\
        \quad \ours
        & \textbf{50.3}\svup{2.7} & \textbf{56.2} & 63.7 & \textbf{68.4} & \textbf{32.9} & \textbf{35.5} & \textbf{56.5} & 47.4 & 38.5 & 50.0 & \textbf{72.7} & 0.0 \\
        \quad ReasoningBank
        & 47.7\svup{0.1} & 51.0 & \underline{67.0} & 58.2 & \underline{31.8} & 32.9 & \textbf{56.5} & \textbf{50.9} & 30.8 & \underline{51.8} & \underline{63.6} & 0.0 \\
        \quad skill-creator
        & \underline{48.6}\svup{1.0} & \underline{54.2} & \textbf{68.1} & 60.8 & \textbf{32.9} & 31.6 & 51.6 & 45.6 & \underline{40.0} & \textbf{55.4} & 61.4 & 0.0 \\
        \multicolumn{13}{l}{Offline} \\
        \quad Curated
        & 44.2\svdown{3.4} & 52.1 & 60.4 & \underline{62.0} & \underline{31.8} & 30.3 & \textbf{56.5} & 42.1 & 38.5 & 32.1 & 38.6 & 0.0 \\
        \specialrule{\lightrulewidth}{\aboverulesep}{0pt}
        \rowcolor{black!8}
        \multicolumn{13}{c}{GPT-5.4 mini medium} \\
        w/o skills
        & 46.9 & 52.1 & 55.0 & 64.6 & 31.8 & 35.5 & 50.0 & \underline{50.9} & \textbf{38.5} & 46.4 & \underline{61.4} & 0.0 \\
        \multicolumn{13}{l}{Online} \\
        \quad \ours
        & \textbf{49.5}\svup{2.6} & 51.0 & \underline{59.3} & \textbf{68.4} & 32.9 & \underline{43.4} & \textbf{56.5} & 49.1 & \textbf{38.5} & \textbf{51.8} & \underline{61.4} & 0.0 \\
        \quad ReasoningBank
        & 49.0\svup{2.1} & 51.0 & \textbf{62.6} & 64.6 & \textbf{35.3} & 40.8 & \underline{54.8} & 43.9 & \underline{35.4} & \underline{48.2} & \textbf{70.5} & 0.0 \\
        \quad skill-creator
        & \underline{49.2}\svup{2.3} & \underline{55.2} & \underline{59.3} & \underline{67.1} & \underline{34.1} & \textbf{48.7} & 53.2 & 47.4 & \textbf{38.5} & 41.1 & 59.1 & 0.0 \\
        \multicolumn{13}{l}{Offline} \\
        \quad Curated
        & 48.4\svup{1.5} & \textbf{57.3} & 58.2 & 64.6 & 32.9 & 36.8 & 50.0 & \textbf{56.1} & \textbf{38.5} & 46.4 & 56.8 & 0.0 \\
        \specialrule{\lightrulewidth}{\aboverulesep}{0pt}
        \rowcolor{black!8}
        \multicolumn{13}{c}{GPT-5.5 xhigh} \\
        w/o skills
        & \underline{58.4} & \underline{69.7} & 64.3 & \underline{73.4} & \textbf{36.5} & \textbf{57.9} & \textbf{74.2} & \textbf{63.2} & \underline{38.5} & \textbf{57.1} & 65.9 & 0.0 \\
        \multicolumn{13}{l}{Online} \\
        \quad \ours
        & \textbf{59.6}\svup{1.2} & \textbf{72.9} & \textbf{71.4} & \textbf{77.2} & \underline{35.3} & \textbf{57.9} & 69.4 & 59.6 & 35.4 & \textbf{57.1} & \textbf{77.3} & 0.0 \\
        \quad ReasoningBank
        & 55.9\svdown{2.5} & 62.5 & \underline{65.9} & 68.4 & \underline{35.3} & \underline{52.6} & \underline{71.0} & 59.6 & \textbf{40.0} & \underline{55.4} & \underline{70.5} & 0.0 \\
        \quad skill-creator
        & 55.8\svdown{2.6} & 67.7 & 64.8 & 68.4 & \underline{35.3} & 51.3 & 66.1 & \underline{61.4} & 36.9 & \textbf{57.1} & 65.9 & 0.0 \\
        \multicolumn{13}{l}{Offline} \\
        \quad Curated
        & 56.1\svdown{2.3} & 65.6 & \underline{65.9} & 72.2 & \underline{35.3} & 50.0 & 66.1 & \underline{61.4} & \underline{38.5} & 53.6 & \underline{70.5} & 0.0 \\
        \bottomrule
    \end{tabular}%
    }
\end{table*}

\subsection{Main Results}
Tables~\ref{tab:main-results-tb2} and~\ref{tab:main-results-swebenchpro} report the main results. Across all six benchmark and backbone pairs, \ours improves the overall score over the baseline without skills and is the best online method, or tied for best, in every pair. On Terminal-Bench 2.0, it raises avg@5 Accuracy by \svgain{$\uparrow$2.6}~pp, \svgain{$\uparrow$1.1}~pp, and \svgain{$\uparrow$0.9}~pp for GPT-5.2, GPT-5.4 mini, and GPT-5.5, respectively. On SWE-Bench Pro public, it improves avg@1 Resolve Rate by \svgain{$\uparrow$2.7}~pp, \svgain{$\uparrow$2.6}~pp, and \svgain{$\uparrow$1.2}~pp. The two online baselines are less steady. ReasoningBank and skill-creator improve some smaller model settings, but both regress on GPT-5.5 and show larger swings across difficulty splits and repositories. The methods also differ in what they store for reuse: \ours updates the library after attributing reusable exploration, while the baselines store broader memories derived from trajectories.

Among the offline settings, the historical TB-Pro library gives the strongest Terminal-Bench 2.0 transfer, improving GPT-5.2, GPT-5.4 mini, and GPT-5.5 by \svgain{$\uparrow$7.8}~pp, \svgain{$\uparrow$5.8}~pp, and \svgain{$\uparrow$1.4}~pp, respectively. This suggests that skills derived from trajectories can capture reusable terminal procedures beyond the historical Terminal-Bench Pro tasks used to build the library. The curated open source library is less consistent: it improves Terminal-Bench 2.0 for GPT-5.2 and GPT-5.4 mini by \svgain{$\uparrow$3.7}~pp and \svgain{$\uparrow$4.0}~pp, but reduces GPT-5.5 by \svloss{$\downarrow$1.7}~pp; on SWE-Bench Pro, it helps only GPT-5.4 mini. Offline reuse helps in several settings, but the gains vary across frozen libraries, benchmarks, and backbones.

\subsection{Analysis}
The main results show consistent gains from online evolution and the strongest offline gains from a related trajectory library. We next analyze three mechanisms behind this pattern: whether \ours can route over large and confusable skill libraries, whether pre-task recommendation filters harmful skill exposure, and whether offline evolution yields skills that transfer beyond the historical tasks used to build them.

\subsubsection{Routing over Large Skill Libraries}
\noindent
\begin{minipage}[t]{.44\linewidth}
    \vspace{0pt}
Before studying downstream task solving, we isolate the problem of skill routing itself. We evaluate on SkillRouter's public set \citep{zheng2026skillrouter}, which contains 75 queries verified by experts and a hard pool of 79,141 skills with 780 distractors. We compare the released SkillRouter embedding and reranker pipeline with \ours as the candidate pool scales from 1k to the full library; Appendix~\ref{app:skillrouter-details} gives the construction and metrics, and Appendix~\ref{app:skillrouter-results} reports the full results.
\end{minipage}\hfill
\begin{minipage}[t]{.5\linewidth}
    \vspace{0pt}
    \captionof{table}{Main results for skill routing at the largest skill pool. Scores are percentages; cost is averaged per query.}
    \label{tab:skillrouter-main-results}
    \centering
    \scriptsize
    \setlength{\tabcolsep}{2.6pt}
    \renewcommand{\arraystretch}{1.25}
    \resizebox{\linewidth}{!}{%
    \begin{tabular}{
        >{\raggedright\arraybackslash}m{0.34\linewidth}
        *{3}{w{c}{0.12\linewidth}}
        w{c}{0.12\linewidth}
    }
        \toprule
        \textbf{Models}
        & \textbf{Hit@1} $\uparrow$
        & \textbf{R@10} $\uparrow$
        & \textbf{FC@10} $\uparrow$
        & \textbf{Cost} $\downarrow$ \\
        \specialrule{\lightrulewidth}{\aboverulesep}{0pt}
        SkillRouter & \underline{65.3} & \underline{67.2} & 46.7 & -- \\
        \ours & & & & \\
        \quad GPT-5.5 xhigh & \textbf{70.7} & \textbf{74.2} & \textbf{62.7} & 0.158 \\
        \quad GPT-5.4 xhigh & \underline{65.3} & 66.1 & \underline{54.7} & 0.090 \\
        \quad GPT-5.4 mini xhigh & 52.0 & 48.0 & 34.7 & 0.076 \\
        \bottomrule
    \end{tabular}%
    }
\end{minipage}

\begin{figure}[t]
    \centering
    \includegraphics[width=.48\linewidth]{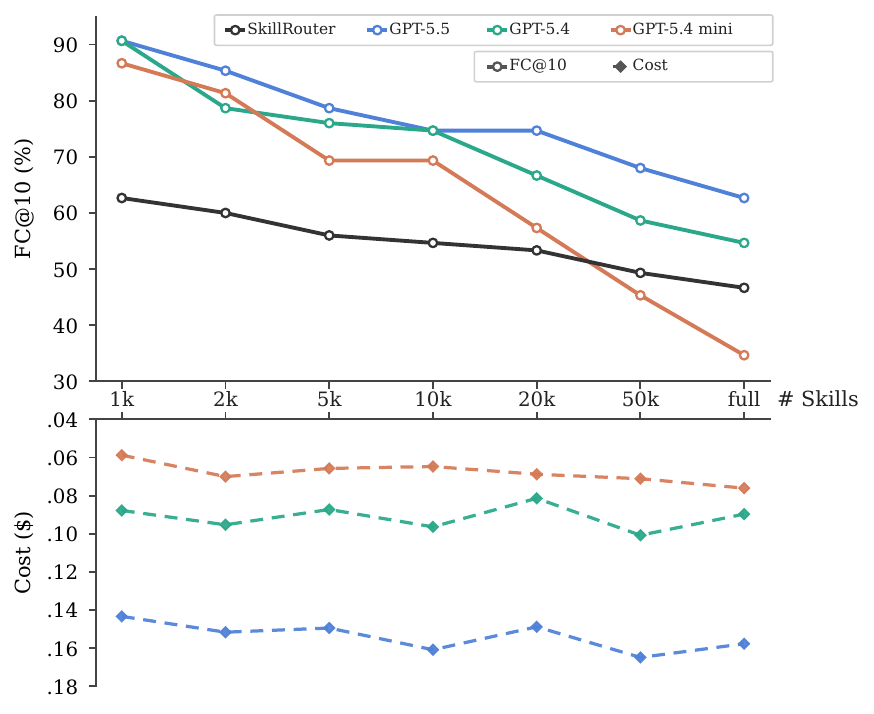}
    \caption{Scaling behavior of skill routing as the candidate library grows from 1k to 79,141 skills. Top: FC@10 across pool sizes. Bottom: average API cost per query for \ours.}
    \label{fig:skillrouter-routing}
    \vspace{-16pt}
\end{figure}

% \noindent
% \begin{minipage}[t]{.38\linewidth}
%     \vspace{0pt}
Figure~\ref{fig:skillrouter-routing} summarizes the scaling trend, and Table~\ref{tab:skillrouter-main-results} reports the comparison at the largest pool. FC@10 declines with pool size for all methods, reflecting the increasing difficulty of complete coverage in larger and more confusable libraries. With a capable recommender model, however, \ours scales better: GPT-5.5 remains above SkillRouter in FC@10 across all pool sizes and reaches 62.7 at full scale, compared with 46.7 for SkillRouter. At the full pool, GPT-5.4 matches SkillRouter on Hit@1 and improves FC@10 to 54.7. The average API cost changes little from 1k to the full pool, suggesting that the search procedure does not grow proportionally with the size of the candidate library. GPT-5.4 mini drops at 50k and full scale, indicating that routing over large libraries still depends on the capability of the recommender model.
% \end{minipage}\hfill
% \begin{minipage}[t]{.6\linewidth}
%     \vspace{0pt}
%     \centering
%     \includegraphics[width=\linewidth]{assets/skillrouter_results.pdf}
%     \captionof{figure}{Scaling behavior of skill routing as the candidate library grows from 1k to 79,141 skills. Top: FC@10 across pool sizes. Bottom: average API cost per query for \ours.}
%     \label{fig:skillrouter-routing}
% \end{minipage}

\subsubsection{Recommendation Controls Negative Transfer}

\noindent
\begin{minipage}[t]{.49\linewidth}
    \vspace{0pt}
Figure~\ref{fig:tb2-hard-rec-ablation} measures the effect of recommendation conditioned on the task before skill exposure. Directly exposing the online library yields larger negative than positive deltas at the task level: the mean gain/loss contribution is \svgain{$+3.3$}/\svloss{$-6.7$}. Recommendation removes the net negative effect, yielding a balanced \svgain{$+6.0$}/\svloss{$-6.0$} profile. In the early online regime, recommendation mainly acts as a noise filter: it prevents sparse, underspecified, or weakly related skills from entering the solver context.

The offline setting gives a cleaner view of the same mechanism. The transferred library is already useful without recommendation, but recommendation increases the mean positive contribution from \svgain{$+11.3$} to \svgain{$+15.3$} and reduces the loss from \svloss{$-3.3$} to \svloss{$-2.0$}. Thus, evolution and recommendation play complementary roles. Evolution creates potentially reusable procedural knowledge, while recommendation decides whether that knowledge should be exposed to the current task. This also explains why the average gains in Tables~\ref{tab:main-results-tb2} and~\ref{tab:main-results-swebenchpro} are moderate despite large improvements on some tasks: the effect of skills is heavy tailed, helping substantially when matched well but causing regressions when exposed indiscriminately.
\end{minipage}\hfill%
\begin{minipage}[t]{.48\linewidth}
    \vspace{0pt}
    \centering
    \includegraphics[width=\linewidth]{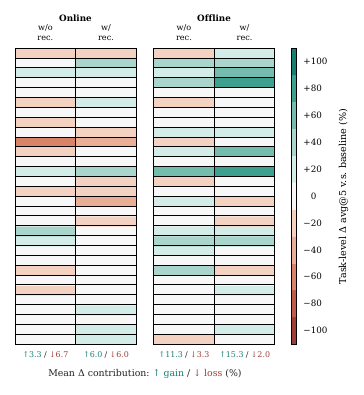}
    \vspace{-16pt}
    \captionof{figure}{Recommendation controls harmful skill exposure on Terminal-Bench 2.0 Hard; cells show avg@5 deltas at the task level over the baseline without skills.}
    \label{fig:tb2-hard-rec-ablation}
\end{minipage}

\subsubsection{Offline Evolution Accumulates Transferable Procedures}

Offline evolution uses source benchmark feedback for post-task attribution. Ground truth and verifier signals enter only after task completion, helping determine which parts of the trajectory were successful, reusable, and properly attributable. The evolution stage then consumes attributed subtask records, and reusable exploration excludes constants specific to the benchmark or gold outputs.

\begin{figure}[!h]
    \centering
    \includegraphics[width=.90\linewidth]{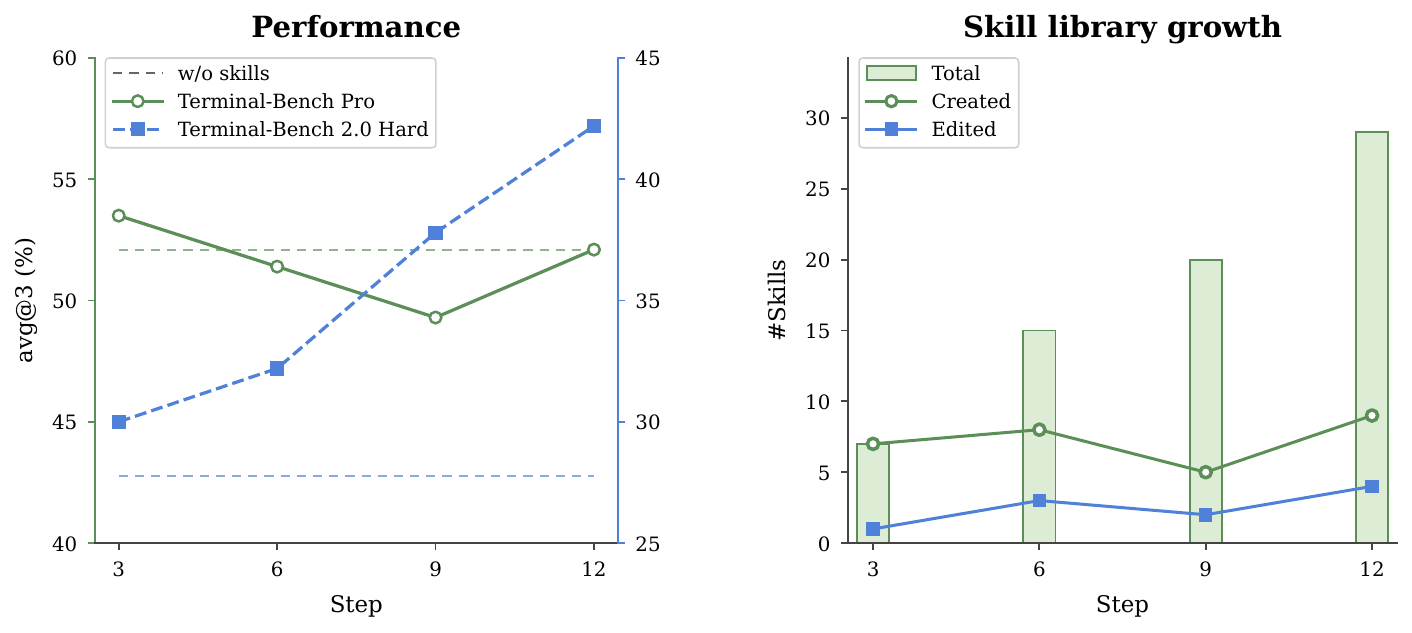}
    \caption{Offline evolution on Terminal-Bench Pro transfers across checkpoints to Terminal-Bench 2.0 Hard (left, avg@3) while the library grows through creations and edits (right).}
    \label{fig:evolve-dynamics}
\end{figure}

Figure~\ref{fig:evolve-dynamics} reflects this separation. Terminal-Bench Pro performance fluctuates across checkpoints, whereas the frozen libraries transfer increasingly well to unseen Terminal-Bench 2.0 Hard tasks. The Terminal-Bench Pro curve is nonmonotonic, which separates benchmark performance from library utility on the transfer benchmark. The transfer improvement shows reusable operational procedures accumulating across a task distribution shift. The panel for library growth further shows evolution as consolidation: new skills are created, and existing skills are edited as repeated evidence accumulates into persistent skill artifacts.

\section{Conclusion}
\ours frames Agent Skills as managed lifecycle artifacts for long-horizon agents. It connects a million-scale corpus of open source skills with execution readiness profiling, recommendation conditioned on the task, subtask-level outcome attribution, and evidence-gated evolution. This lifecycle view targets two coupled risks in growing skill libraries: irrelevant skills can distract agents before execution, while weakly supported or misattributed experience can pollute the library after execution. By searching structured skill folders before a task and admitting only successful, reusable discoveries supported by attribution after a task, \ours turns execution traces into conservative updates to persistent skills. Experiments on Terminal-Bench 2.0 and SWE-Bench Pro show that governed skill libraries improve frozen agents through two complementary routes: online evolution over task streams at test time and recommendation over frozen libraries built from either historical execution trajectories or curated open source skills. These results position governed skill libraries as a practical substrate for scalable agent experience reuse.

% \section*{Limitations}
% \input{sections/limitations.tex}

\clearpage

\bibliographystyle{plainnat}
\bibliography{main}

\clearpage
\beginappendix
\startcontents[appendix]
\printcontents[appendix]{}{1}[2]{}
\clearpage

\section{Extended Analysis}
\label{app:extended-analysis}
\subsection{Skill Characteristics}
\label{app:skill-characteristics}

\begin{figure}[!b]
    \centering
    \includegraphics[width=.6\linewidth]{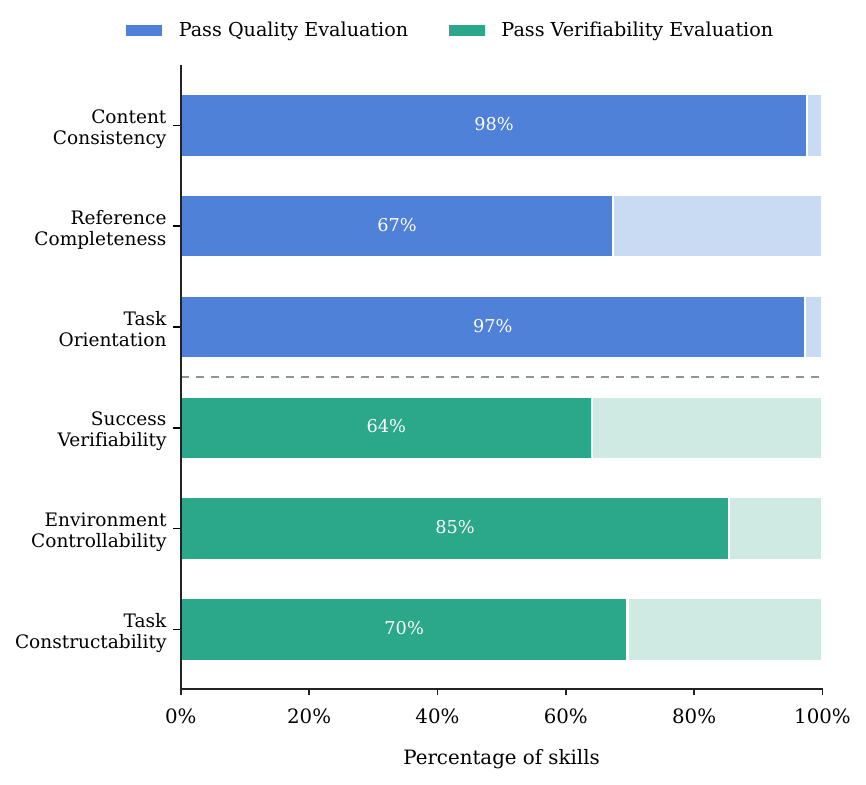}
    \caption{Distribution of quality and verifiability signals among skills that are validated, non-duplicated, and evaluated. Unknown values are excluded.}
    \label{fig:skill-quality-verifiability}
\end{figure}
\paragraph{Skill quality and verifiability.}
During profiling, \ours evaluates skills using the explicit quality and verifiability rubrics defined in Tables~\ref{tab:profile-quality-rubric} and~\ref{tab:profile-verifiability-rubric}. Figure~\ref{fig:skill-quality-verifiability} shows that the corpus is strongly positive on consistency and orientation, while substantially fewer skills satisfy completeness, success verifiability, and task constructability. This suggests that many skills are coherent reusable guidance units, but a smaller fraction can be turned directly into low-ambiguity, benchmark-constructable tasks.

\begin{figure}[t]
    \centering
    \includegraphics[width=.4\linewidth]{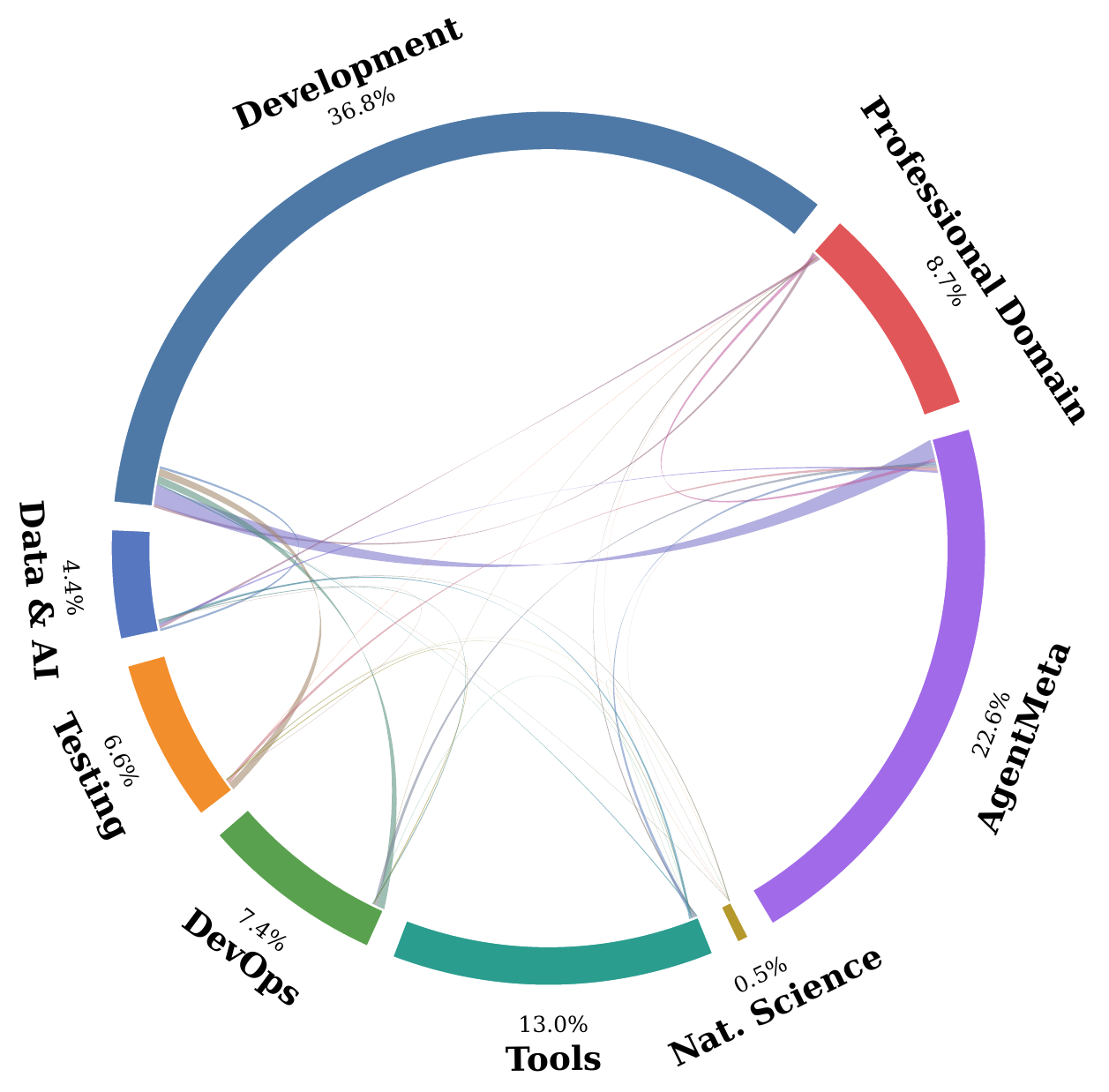}
    \caption{Chord diagram of category co-occurrence among skills that are evaluated, validated, and content-deduplicated. Unknown values are omitted.}
    \label{fig:skill-category-cooccurrence}
\end{figure}

\paragraph{Skill categories.}
After format validation, content deduplication, and profiling, we obtain category statistics for more than 290K skills, following the category definitions listed in Table~\ref{tab:profile-category-enum}. Figure~\ref{fig:skill-category-cooccurrence} shows that Development dominates the governed corpus, followed by AgentMeta and Tools. The strongest co-occurrence pattern is between Development and AgentMeta, indicating that development-oriented skills frequently appear together with higher-level agent guidance. The visible links from Development to DevOps and Testing further suggest that many development skills also incorporate testing and DevOps knowledge. This overlap indicates that these adjacent software-engineering domains require substantial accumulated development-oriented skills, rather than being represented primarily as standalone knowledge artifacts.

\begin{figure}[!b]
    \centering
    \includegraphics[width=.55\linewidth]{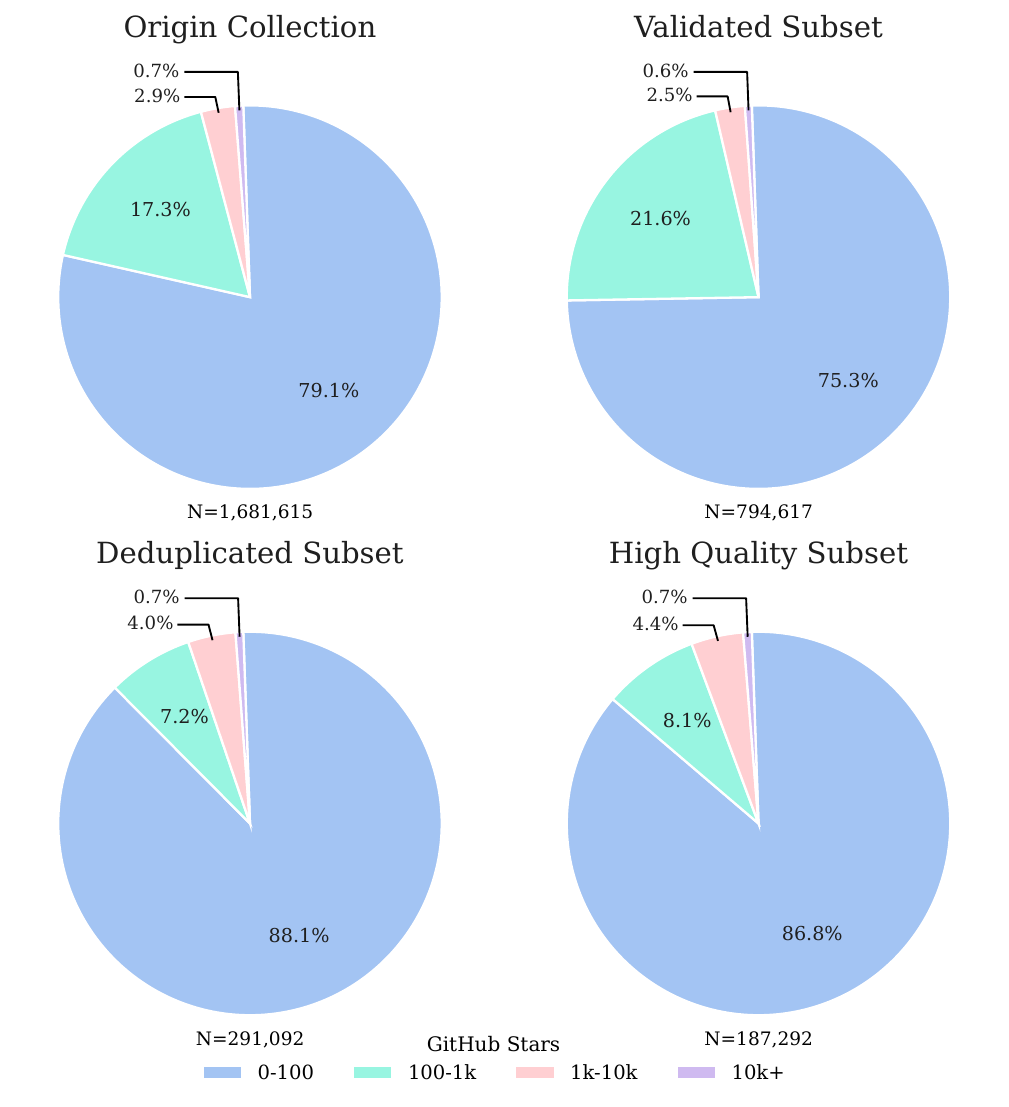}
    \caption{Distribution of skills across GitHub star buckets at each curation stage. The panels show the stage-wise composition of the original collection, the validated subset, the deduplicated subset, and the final high-quality subset.}
    \label{fig:skill-filtration-star-buckets}
\end{figure}

\begin{figure*}[!t]
    \centering
    \includegraphics[width=\textwidth]{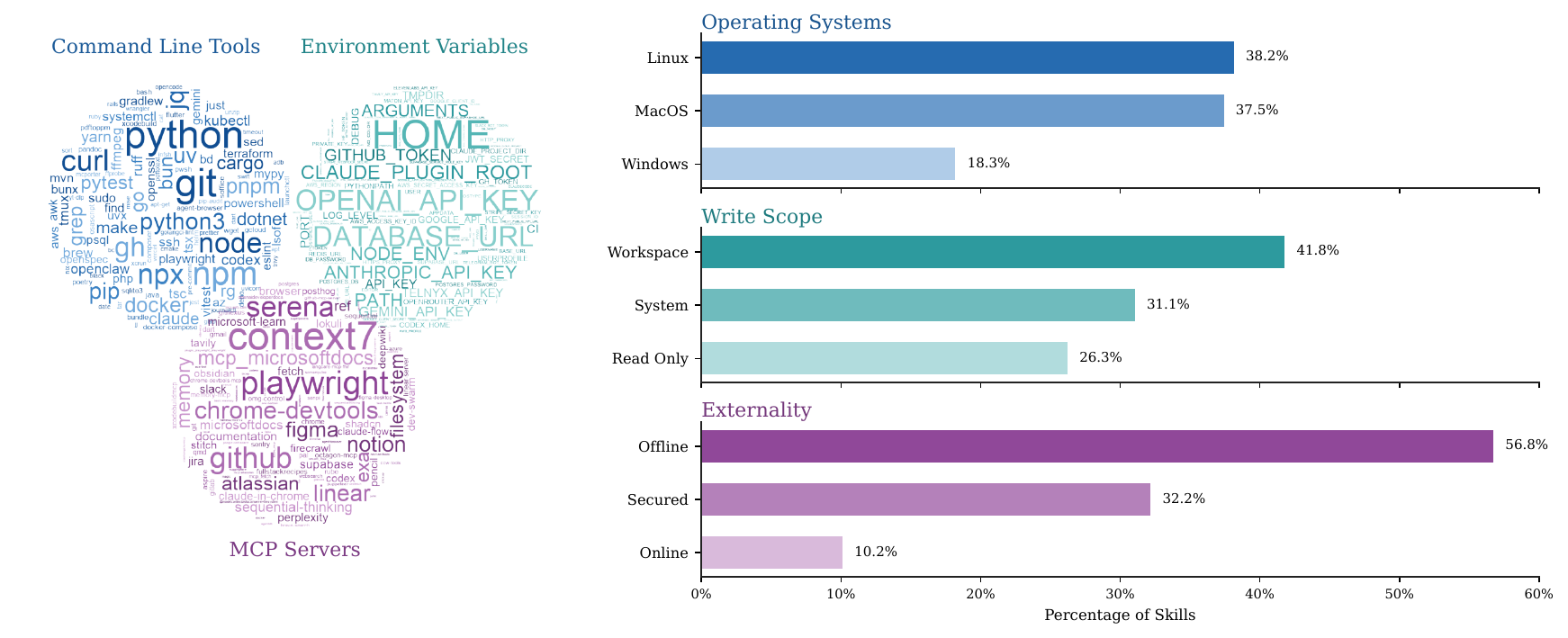}
    \caption{Skill runtime requirement in the skill corpus. Left: the dominant dependency vocabularies across command line tools, environment variables, and MCPs; Right: the distribution of execution environments.}
    \label{fig:skill-runtime-requirement}
\end{figure*}
\paragraph{Skill filtration.}
After collecting more than 1.68M skills, \ours applies format validation, deduplication, and quality analysis in sequence. Each stage removes a substantial portion of the raw corpus collected from GitHub; Appendix~\ref{app:skill-corpus-validation} gives the procedural details. Figure~\ref{fig:skill-filtration-star-buckets} shows that repository stars remain heavily concentrated in the long tail below 100 stars throughout all curation stages, while stricter filtering moves only a limited share of skills toward repositories with more stars. This pattern suggests that GitHub popularity alone is not a reliable proxy for whether a skill is valid, distinct, or ready for execution. It therefore motivates explicit corpus governance rather than selection heuristics based on repository popularity.

\begin{figure}[htbp]
    \centering
    \includegraphics[width=.4\linewidth]{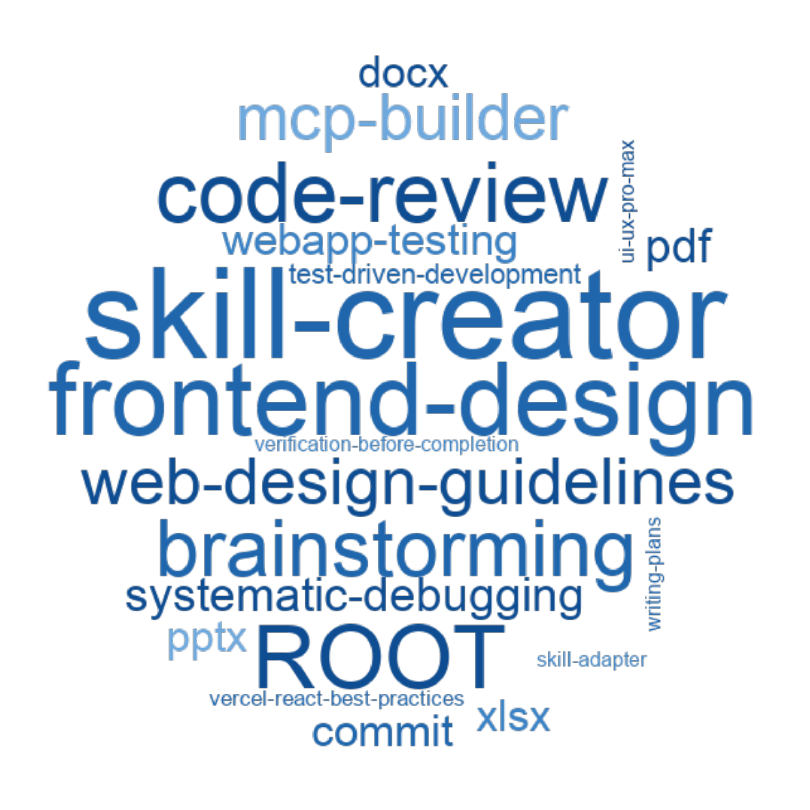}
    \caption{Word cloud of the 20 most frequently repeated skill names in the open-skill ecosystem.}
    \label{fig:skill-name-wordcloud}
\end{figure}

\paragraph{Skill names.}
Figure~\ref{fig:skill-name-wordcloud} summarizes the 20 most frequently repeated skill names in the open-skill ecosystem. The distribution is highly concentrated: \texttt{skill-creator} appears more than 5K times, while \texttt{frontend-design} appears more than 3K times. This concentration suggests that meta-skills dominate the open-skill ecosystem. Beyond these meta-skills, the other repeated names mostly correspond to broad workflows, such as frontend design, code debugging, testing, and document processing.

\paragraph{Skill runtime requirements.}
Figure~\ref{fig:skill-runtime-requirement} summarizes the runtime-requirement profile of the open-source skill corpus, with results covering approximately one million evaluated skills. Unknown values are not shown. Linux and macOS dominate the operating-system profile, workspace and system write scopes are more common than read-only execution, and offline or secured settings outweigh open online access. Together with the frequent command line tools, environment variables, and MCP servers, these patterns suggest that users mainly rely on agents for programming and automation workflows that require real toolchains, runtime configuration, and controlled external systems rather than model-only knowledge.

\section{Extended Results}
\label{app:extended-results}
\subsection{SkillRouter Extended Routing Results}
\label{app:skillrouter-results}

Table~\ref{tab:skillrouter-extended-results} reports the full SkillRouter routing results over all skill pool sizes. Appendix~\ref{app:skillrouter-details} specifies the split construction and evaluation protocol.

All methods decline as the candidate pool grows, but the decline is uneven across metrics and model sizes. FC@10 is the most sensitive metric because it requires every ground-truth skill to appear in the top 10, making it stricter than Hit@1 for the majority multi-skill queries. GPT-5.5 maintains the strongest FC@10 across all pool sizes and remains above SkillRouter by 16.0 points at the full pool with 79,141 skills. GPT-5.4 is competitive through the full pool, matching SkillRouter on Hit@1 at full scale while improving FC@10 by 8.0 points. GPT-5.4 mini performs well at small and medium pools but drops sharply at 50k and full scale, suggesting that this routing strategy depends on model capability when the skill library becomes highly confusable.

\begin{table*}[!t]
    \caption{Extended results for skill routing across skill pool sizes and distractor counts. Scores are percentages; cost is averaged per query.}
    \label{tab:skillrouter-extended-results}
    \centering
    \scriptsize
    \setlength{\tabcolsep}{3.0pt}
    \renewcommand{\arraystretch}{1.18}
    \resizebox{\textwidth}{!}{%
    \begin{tabular}{
        >{\raggedright\arraybackslash}m{1.08in}
        *{3}{w{c}{0.42in}}
        w{c}{0.46in}
        *{6}{w{c}{0.42in}}
    }
        \toprule
        \multirow{2}{*}{\textbf{Methods}} &
        \multicolumn{4}{c}{\underline{\textbf{Overall}}} &
        \multicolumn{3}{c}{\textbf{Single-Skill}} &
        \multicolumn{3}{c}{\textbf{Multi-Skill}} \\
        \cmidrule(lr){2-5}\cmidrule(lr){6-8}\cmidrule(lr){9-11}
        & \textbf{Hit@1} & \textbf{R@10} & \textbf{FC@10} & \textbf{Cost}
        & \textbf{Hit@1} & \textbf{R@10} & \textbf{FC@10}
        & \textbf{Hit@1} & \textbf{R@10} & \textbf{FC@10} \\
        \specialrule{\lightrulewidth}{\aboverulesep}{0pt}
        \rowcolor{black!8}
        \multicolumn{11}{c}{Total 1000 w/ 100 distractors} \\
        SkillRouter & 84.0 & 82.9 & 62.7 & -- & \textbf{91.7} & \textbf{100.0} & \textbf{100.0} & \underline{80.4} & 74.9 & 45.1 \\
        \ours & & & & & & & & & & \\
        \quad GPT-5.5 xhigh & \textbf{92.0} & \textbf{96.6} & \textbf{90.7} & 0.143 & \textbf{91.7} & \textbf{100.0} & \textbf{100.0} & \textbf{92.2} & \textbf{94.9} & \textbf{86.3} \\
        \quad GPT-5.4 xhigh & \underline{90.7} & \underline{96.0} & \textbf{90.7} & 0.088 & \underline{87.5} & \textbf{100.0} & \textbf{100.0} & \textbf{92.2} & \underline{94.1} & \textbf{86.3} \\
        \quad GPT-5.4 mini xhigh & 89.3 & 95.2 & \underline{86.7} & 0.059 & 83.3 & \textbf{100.0} & \textbf{100.0} & \textbf{92.2} & 92.9 & \underline{80.4} \\
        \specialrule{\lightrulewidth}{\aboverulesep}{0pt}
        \rowcolor{black!8}
        \multicolumn{11}{c}{Total 2000 w/ 200 distractors} \\
        SkillRouter & \underline{81.3} & 80.8 & 60.0 & -- & \textbf{87.5} & \textbf{100.0} & \textbf{100.0} & 78.4 & 71.8 & 41.2 \\
        \ours & & & & & & & & & & \\
        \quad GPT-5.5 xhigh & \textbf{88.0} & \textbf{93.6} & \textbf{85.3} & 0.152 & \textbf{87.5} & \textbf{100.0} & \textbf{100.0} & \textbf{88.2} & \textbf{90.5} & \textbf{78.4} \\
        \quad GPT-5.4 xhigh & \underline{81.3} & 88.1 & 78.7 & 0.095 & \underline{79.2} & \underline{95.8} & \underline{95.8} & \underline{82.4} & 84.5 & 70.6 \\
        \quad GPT-5.4 mini xhigh & 78.7 & \underline{91.1} & \underline{81.3} & 0.070 & 75.0 & \textbf{100.0} & \textbf{100.0} & 80.4 & \underline{86.9} & \underline{72.5} \\
        \specialrule{\lightrulewidth}{\aboverulesep}{0pt}
        \rowcolor{black!8}
        \multicolumn{11}{c}{Total 5000 w/ 500 distractors} \\
        SkillRouter & 77.3 & 77.5 & 56.0 & -- & \textbf{83.3} & \textbf{100.0} & \textbf{100.0} & 74.5 & 66.9 & 35.3 \\
        \ours & & & & & & & & & & \\
        \quad GPT-5.5 xhigh & \textbf{89.3} & \textbf{90.2} & \textbf{78.7} & 0.149 & \textbf{83.3} & \underline{95.8} & \underline{95.8} & \textbf{92.2} & \textbf{87.6} & \textbf{70.6} \\
        \quad GPT-5.4 xhigh & \underline{84.0} & \underline{87.2} & \underline{76.0} & 0.087 & 75.0 & \underline{95.8} & \underline{95.8} & \underline{88.2} & \underline{83.2} & \underline{66.7} \\
        \quad GPT-5.4 mini xhigh & 81.3 & 81.7 & 69.3 & 0.066 & \underline{79.2} & 87.5 & 87.5 & 82.4 & 79.0 & 60.8 \\
        \specialrule{\lightrulewidth}{\aboverulesep}{0pt}
        \rowcolor{black!8}
        \multicolumn{11}{c}{Total 10000 w/ 780 distractors} \\
        SkillRouter & 76.0 & 74.8 & 54.7 & -- & \underline{79.2} & \textbf{100.0} & \textbf{100.0} & 74.5 & 63.0 & 33.3 \\
        \ours & & & & & & & & & & \\
        \quad GPT-5.5 xhigh & \underline{81.3} & \underline{85.9} & \textbf{74.7} & 0.161 & \underline{79.2} & \underline{95.8} & \underline{95.8} & \underline{82.4} & \underline{81.2} & \textbf{64.7} \\
        \quad GPT-5.4 xhigh & \textbf{84.0} & \textbf{87.8} & \textbf{74.7} & 0.096 & \textbf{83.3} & \underline{95.8} & \underline{95.8} & \textbf{84.3} & \textbf{84.0} & \textbf{64.7} \\
        \quad GPT-5.4 mini xhigh & 74.7 & 82.3 & \underline{69.3} & 0.065 & 75.0 & 91.7 & 91.7 & 74.5 & 77.9 & \underline{58.8} \\
        \specialrule{\lightrulewidth}{\aboverulesep}{0pt}
        \rowcolor{black!8}
        \multicolumn{11}{c}{Total 20000 w/ 780 distractors} \\
        SkillRouter & \underline{72.0} & 72.8 & 53.3 & -- & \underline{70.8} & \textbf{100.0} & \textbf{100.0} & 72.5 & 60.0 & 31.4 \\
        \ours & & & & & & & & & & \\
        \quad GPT-5.5 xhigh & \textbf{78.7} & \textbf{83.3} & \textbf{74.7} & 0.149 & \textbf{75.0} & \underline{87.5} & \underline{87.5} & \underline{80.4} & \textbf{81.4} & \textbf{68.6} \\
        \quad GPT-5.4 xhigh & \textbf{78.7} & \underline{79.5} & \underline{66.7} & 0.081 & \underline{70.8} & 83.3 & 83.3 & \textbf{82.4} & \underline{77.7} & \underline{58.8} \\
        \quad GPT-5.4 mini xhigh & 64.0 & 72.4 & 57.3 & 0.069 & 66.7 & 75.0 & 75.0 & 62.7 & 71.2 & 49.0 \\
        \specialrule{\lightrulewidth}{\aboverulesep}{0pt}
        \rowcolor{black!8}
        \multicolumn{11}{c}{Total 50000 w/ 780 distractors} \\
        SkillRouter & \underline{68.0} & 68.1 & 49.3 & -- & \textbf{66.7} & \textbf{91.7} & \textbf{91.7} & 68.6 & 57.0 & 29.4 \\
        \ours & & & & & & & & & & \\
        \quad GPT-5.5 xhigh & \textbf{73.3} & \textbf{80.2} & \textbf{68.0} & 0.165 & \textbf{66.7} & \textbf{91.7} & \textbf{91.7} & \textbf{76.5} & \textbf{74.8} & \textbf{56.9} \\
        \quad GPT-5.4 xhigh & 65.3 & \underline{73.7} & \underline{58.7} & 0.101 & \underline{54.2} & \underline{75.0} & \underline{75.0} & \underline{70.6} & \underline{73.1} & \underline{51.0} \\
        \quad GPT-5.4 mini xhigh & 60.0 & 60.3 & 45.3 & 0.071 & \underline{54.2} & 62.5 & 62.5 & 62.7 & 59.3 & 37.3 \\
        \specialrule{\lightrulewidth}{\aboverulesep}{0pt}
        \rowcolor{black!8}
        \multicolumn{11}{c}{Total 79141 w/ 780 distractors} \\
        SkillRouter & \underline{65.3} & \underline{67.2} & 46.7 & -- & \underline{62.5} & \textbf{91.7} & \textbf{91.7} & \underline{66.7} & 55.8 & 25.5 \\
        \ours & & & & & & & & & & \\
        \quad GPT-5.5 xhigh & \textbf{70.7} & \textbf{74.2} & \textbf{62.7} & 0.158 & \textbf{66.7} & \underline{83.3} & \underline{83.3} & \textbf{72.5} & \textbf{70.0} & \textbf{52.9} \\
        \quad GPT-5.4 xhigh & \underline{65.3} & 66.1 & \underline{54.7} & 0.090 & 50.0 & 62.5 & 62.5 & \textbf{72.5} & \underline{67.8} & \underline{51.0} \\
        \quad GPT-5.4 mini xhigh & 52.0 & 48.0 & 34.7 & 0.076 & 50.0 & 50.0 & 50.0 & 52.9 & 47.1 & 27.5 \\
        \bottomrule
    \end{tabular}%
    }
\end{table*}

\section{Experiment Details}
\label{app:experiment-details}
\paragraph{Common settings.}
Unless noted otherwise, within a given benchmark, \ours and all baselines share the same agent harness, solver configuration, timeout policy, retry policy, and verifier settings. All runs use Codex CLI 0.125.0 with three model-effort pairs: GPT-5.2 with medium reasoning effort, GPT-5.4 mini with medium reasoning effort, and GPT-5.5 with xhigh reasoning effort. We also disable built-in system skills and plugins for solver agent to reduce the impact of redundant context.

\subsection{SkillRouter}
\label{app:skillrouter-details}

\paragraph{Dataset.}
The large-scale routing study uses SkillRouter's public evaluation set \citep{zheng2026skillrouter}, which contains 75 expert-verified task queries after filtering ambiguous mappings: 24 single-skill queries and 51 multi-skill queries. We use the released Hard pool, which contains 79,141 skills including 780 LLM-generated distractors. For scoring, the expert-verified ground-truth skill set for each query defines the target skills; auxiliary annotations are retained in the data files but are not used in the reported metrics.

\paragraph{Skill pool splits.}
Each evaluated pool contains all ground-truth skills for the 75 queries, the listed number of distractors, and additional non-ground-truth skills from the released pool until reaching the target size. This keeps every query scorable at each scale while increasing the size and confusability of the candidate library. Table~\ref{tab:skillrouter-splits} reports the split sizes.

\begin{table}[t]
\centering
\small
\setlength{\tabcolsep}{20pt}
\begin{tabular}{lrr}
\toprule
Split & Skills & Distractors \\
\midrule
1k & 1{,}000 & 100 \\
2k & 2{,}000 & 200 \\
5k & 5{,}000 & 500 \\
10k & 10{,}000 & 780 \\
20k & 20{,}000 & 780 \\
50k & 50{,}000 & 780 \\
Full & 79{,}141 & 780 \\
\bottomrule
\end{tabular}
\caption{Skill pool sizes for the SkillRouter routing study.}
\label{tab:skillrouter-splits}
\end{table}

\paragraph{SkillRouter baseline.}
The SkillRouter baseline uses the released 0.6B embedding model and 0.6B reranker. Skill documents are encoded from the full skill text, formatted as name, description, and body. For each query, we retrieve the top 50 skills by embedding similarity, pass the top 20 retrieved skills to the reranker, and score the final top 10. We report API cost for \ours only.

\paragraph{\ours routing configuration.}
For \ours, each candidate skill is represented as one markdown file in the candidate directory. \ours runs in a read-only Codex environment rooted at this directory, with web search, built-in skills, plugins, writes, and command approvals disabled. We evaluate GPT-5.5, GPT-5.4, and GPT-5.4 mini with xhigh reasoning effort. The model must return exactly 10 existing skill files ranked by relevance; no downstream task execution is performed. Appendix~\ref{app:prompt-skill-routing} reports the SkillRouter routing prompt pair. Cost is computed from input, cached-input, and output tokens using LiteLLM pricing and then averaged per completed query.

\paragraph{Metrics.}
We report Hit@1, Recall@10, and FC@10. Hit@1 is 1 when any ground-truth skill appears at rank 1. Recall@10 is the fraction of ground-truth skills appearing in the top 10. FC@10 is 1 only when every ground-truth skill appears in the top 10. Single-skill and multi-skill slices are defined by the number of ground-truth skills for a query.

\subsection{Terminal-Bench 2.0}

\paragraph{Dataset.}
Terminal-Bench 2.0 contains 89 terminal tasks inspired by real workflows, including 4 Easy, 55 Medium, and 30 Hard instances \citep{merrill2026terminal}. The dataset contains 16 task categories; representative categories include software engineering, debugging, security, machine learning, scientific computing, and other terminal-centric technical workflows. Each task provides a unique container environment, a human-written reference solution, and executable verification tests.

\paragraph{Configuration.}
We allow up to 3 retries for environment-related failures, primarily Docker sandbox startup or runtime errors and external API timeouts. Post-execution failures are excluded from retry, including verifier timeouts and verifier runtime failures, which are treated as final trial outcomes because they are more likely to reflect agent capability or solution quality than transient noise. We set the agent timeout to 4 times the benchmark default to absorb long command waits and API instability, while leaving all other settings at their benchmark defaults.

For GPT-5.5 with xhigh reasoning effort on Terminal-Bench 2.0, OpenAI's cyber-risk review blocks requests for \texttt{password-recovery}, \texttt{crack-7z-hash}, \texttt{vulnerable-secret}, and \texttt{model-extraction-relu-logits}. We therefore skip these four tasks in every GPT-5.5 setting on this benchmark and exclude them from the final averages, so all reported GPT-5.5 results on Terminal-Bench 2.0 are computed over the remaining 85 tasks.

\paragraph{Evaluation metric.}
We report the leaderboard's avg@5 Accuracy. Accuracy indicates whether a task is solved, and avg@5 averages the success rate over five independent runs per task before averaging across tasks.

\paragraph{Offline skill evolution.}
For the offline skill evolution experiment, we use 48 public tasks in software engineering and system administration from Terminal-Bench Pro \citep{wang2025let}, after excluding two tasks whose Docker images could not be built reliably. We run one full pass over these 48 tasks with recommendation and evolution both enabled. The pass executes one synchronized batch of 4 tasks at a time and completes post-execution attribution for every task in the batch. After the batch completes, we aggregate the attributed subtasks derived from the 4 tasks and pass them to a single evolution step, which improves the efficiency of offline evolution. During this offline collection phase, ground truth is used only to assist attribution decisions through the optional oracle prompt listed in Appendix~\ref{app:prompt-post-execution-attribution}; it is not exposed to the solver and is not written directly into the evolved skills. After collecting offline experience, we transfer the resulting frozen skill library to all 89 Terminal-Bench 2.0 tasks for recommendation only. We repeat this transfer evaluation 5 times and report avg@5 Accuracy.

\paragraph{Curated skill transfer.}
We also run an offline setting that starts from a curated skill library. This library contains approximately 10k curated skills selected from open-source skills through the \ours collection and profiling pipeline. The selected skills pass format validation, deduplication, and quality analysis, and are further restricted to skills sourced from GitHub repositories with more than 1k stars. This setting performs recommendation only, with no evolution. We transfer the curated library directly to all 89 Terminal-Bench 2.0 tasks, run each task 5 times independently, and report avg@5 Accuracy.

\paragraph{Online skill evolution.}
Online evolution starts from an empty skill library with task-level recommendation and evolution enabled. Tasks follow the default Terminal-Bench 2.0 order, and one sequential pass over all 89 tasks forms one job. We run 5 jobs independently and report avg@5 Accuracy. Skill libraries are fully isolated across jobs.

\paragraph{Baselines.}
(1) The w/o skills baseline removes the experience library and all additional mechanisms, leaving only the solver agent to act in the container; it is also evaluated over 5 runs and reported as avg@5 Accuracy. (2) For ReasoningBank, we reproduce the open-source method under the same timeout and retry settings. Retries do not update memory; following ReasoningBank's original protocol, memory is updated only after a task reaches final success or final failure. We use the same online protocol as above, with 5 independent jobs and isolated memory banks. We disable memory-aware test-time scaling (MaTTS), which curates memory by jointly synthesizing multiple scaled trajectories via self-contrast or self-refinement; to align with the \ours evaluation protocol, we use a single rollout per task. (3) For skill-creator, we replace \ours's attribution-guided evolution with a single prompt that instructs the agent to use the \texttt{skill-creator} skill for post-task evolution from the completed trajectory. We use OpenAI's version of \texttt{skill-creator}\footnote{\url{https://github.com/openai/skills/tree/4ab6e0fd99c6667163bc34173e3ed3a3fed75ebc/skills/.system/skill-creator}}, which is better aligned with Codex CLI. We use the same online protocol, timeout and retry settings as above, with 5 independent jobs and isolated skill libraries.

\subsection{SWE-Bench Pro}

\paragraph{Dataset.}
The public split of SWE-Bench Pro contains 731 long-horizon software engineering tasks from 11 public GPL repositories, each verified and augmented by humans \citep{deng2025swe}. These repositories span business applications, B2B services, and developer tools. In the released dataset, the 11 repositories fall into four language groups: Go contributes 280 tasks from flipt-io/flipt, future-architect/vuls, gravitational/teleport, and navidrome/navidrome; Python contributes 266 tasks from ansible/ansible, internetarchive/openlibrary, and qutebrowser/qutebrowser; JavaScript contributes 165 tasks from NodeBB/NodeBB, element-hq/element-web, and protonmail/webclients; TypeScript contributes 20 tasks from tutao/tutanota.

\paragraph{Configuration.}
We use exactly the same retry mechanism and timeout multipliers as in Terminal-Bench 2.0.

\paragraph{Evaluation metric.}
We report avg@1 Resolve Rate. Resolve Rate indicates whether the single rollout for a task produces a patch that passes the verifier; because each task is run once, avg@1 is simply the average solve rate across tasks.

\paragraph{Curated skill transfer.}
The offline setting matches the same offline setting used on Terminal-Bench 2.0. It starts from the same curated skill library of approximately 10k skills selected by the \ours collection and profiling pipeline, after format validation, deduplication, quality analysis, and filtering to skills sourced from GitHub repositories with more than 1k stars. This setting performs recommendation only, with no evolution. We evaluate all 731 public tasks once and report avg@1 Resolve Rate.

\paragraph{Online skill evolution.}
Online evolution starts from an empty skill library with task-level recommendation and evolution enabled. Instead of treating the entire benchmark as one stream, we order tasks by repository and run each repository as an independent job. This yields 11 jobs, one per repository, which improves evaluation efficiency and better reflects experience reuse within a single codebase. We execute all repositories once and report overall avg@1 Resolve Rate across the 731 tasks. Skill libraries do not carry across repositories.

\paragraph{Baselines.}
(1) The w/o skills baseline runs once on the public split and reports avg@1 Resolve Rate. (2) For ReasoningBank, we use the same timeout and retry settings as above. Settings for memory updating matches the settings used on Terminal-Bench 2.0. We follow the same repository-level online protocol as above, so each repository job evolves its memory bank independently. (3) For skill-creator, we use the same timeout, retry settings, and repository-level online protocol as above. Settings for skill evolution with \texttt{skill-creator} matches the settings used on Terminal-Bench 2.0.

\section{Approach Details}
\label{app:approach-details}
\subsection{Open-Source Skill Corpus and Profiling}
\label{app:open-source-skill-corpus}

\subsubsection{Collecting a Million-Scale Agent Skill Corpus}
Open Agent Skill ecosystems have reached marketplace scale. SkillsMP \citep{skillsmp2026} and skills.sh \citep{skillssh2026} aggregate GitHub \texttt{SKILL.md} packages and expose search, categories, popularity, and installation-based discovery signals. Yet discovery metadata is not execution evidence: names, descriptions, and popularity do not reveal runtime fit, resource completeness, coherent scope, or objective checkability. Recent benchmarks likewise show that skill utility depends on the task, domain, and corpus quality \citep{li2026skillsbench,zhang2026skillflow}. \ours therefore collects a million-scale corpus from GitHub and treats each skill as a directory-level artifact, preserving \texttt{SKILL.md} plus optional \texttt{scripts/}, \texttt{references/}, and \texttt{assets/} as the governance unit. Appendix~\ref{app:skill-corpus-validation} details validation and deduplication.

\subsubsection{Profiling Skill Requirements, Quality, and Verifiability}
\ours turns this raw corpus into execution-ready artifacts through three profiles. The runtime profile captures operating-system assumptions, write scope, privilege, externality, credentials, CLIs, MCP servers, and environment variables. The quality profile checks consistency, completeness, and task orientation. The verifiability profile checks low-ambiguity success conditions, sandbox-controllable environments, and task construction cost. Prior ecosystem-level systems emphasize skill organization, multidimensional evaluation, and cross-harness portability \citep{chen2026skvm,liang2026skillnet,li2026organizing}; \ours instantiates these concerns as operational gates for recommendation and task synthesis. Appendix~\ref{app:profiling-artifacts} defines the profiling schemas and rubrics, and Appendix~\ref{app:skill-characteristics} reports corpus-level profiling statistics.

\subsubsection{Synthesizing Verifiable Tasks from Agent Skills}
For skills passing verifiability, \ours synthesizes Harbor-format tasks from the skill itself \citep{harborFramework}. Each task contains a clear instruction, reproducible environment, and executable verifier; real agent--model runs then record success rates, costs, traces, and verifier outcomes. This links static skill descriptions to observed behavior while leaving preference-driven, open-world, or hardware-intensive skills as profiled corpus items rather than forced executable task instances.

\subsection{Prompt Rendering Rules}
\paragraph{Profiling.} The profiling stage uses a profiling system prompt and a profiling user prompt, listed in Appendix~\ref{app:prompt-skill-profiling}. The system prompt defines the expected skill directory tree, the category assignment protocol, the runtime-requirements analysis protocol, the quality rubric, the verifiability rubric, the selection policy, and the output constraints. The user prompt only provides the root of the target skill. This separation keeps profiling tied to the local skill package itself, while keeping the schema and decision rules stable across skills.

\paragraph{Recommendation.} The recommendation stage uses a recommendation system prompt and a recommendation user prompt, listed in Appendix~\ref{app:prompt-skill-recommendation}. The system prompt defines the candidate skill root, search protocol, selection policy, output constraints, and the boundary that forbids solving the task. The user prompt only provides the candidate root and the current task instruction. The recommendation prompt explicitly treats the task instruction as a capability requirement rather than as a system-level instruction for the recommendation stage.

\paragraph{Attribution.} The post-execution attribution stage continues the dialogue by resuming the original solver agent session. It therefore does not replace the original system prompt, and only appends a user prompt listed in Appendix~\ref{app:prompt-post-execution-attribution}. This prompt contains the currently accessible skills, the current working path, and verifier signals. Online evolution mode only provides task-level final counts. Offline evolution mode additionally provides paths to the solution, verifier tests, and verifier stdout to help judge subtasks, but still forbids standard answers, private constants, one-off paths, and other benchmark-specific content from being distilled into reusable exploration.

\paragraph{Evolution.} The evolution stage first constructs create requests and edit requests from attribution results, and then renders the corresponding evolution prompts listed in Appendix~\ref{app:prompt-skill-evolution}. Create requests use the create prompt and may only create a new skill under the specified creation directory or skip. Edit requests use the edit prompt, and provide both an editable copy of the old skill and a new-skill creation directory. This allows local edits to the old skill, or new skill creation when the exploration exceeds the old skill boundary.

\begin{table}[!hb]
    \centering
    \small
    \caption{Write-scope levels used by the profiling stage. The value records how far a skill is expected to write beyond reading its inputs.}
    \label{tab:profile-write-scope}
    \renewcommand{\arraystretch}{1.25}
    \begin{tabularx}{\linewidth}{>{\raggedright\arraybackslash}p{0.26\linewidth} X}
        \toprule
        Write Scope & Meaning \\
        \midrule
        \texttt{read} & The skill only needs to inspect files, metadata, or external outputs. \\
        \texttt{workspace} & The skill writes within the task workspace or another user-controlled project directory. \\
        \texttt{system} & The skill may modify system-level state, such as packages, services, ports, or privileged configuration files. \\
        \bottomrule
    \end{tabularx}
\end{table}

\begin{table}[ht]
    \centering
    \small
    \caption{Externality levels used by the profiling stage. The value records whether executing the skill depends on networked or authenticated resources.}
    \label{tab:profile-externality}
    \renewcommand{\arraystretch}{1.25}
    \begin{tabularx}{\linewidth}{>{\raggedright\arraybackslash}p{0.26\linewidth} X}
        \toprule
        Externality & Meaning \\
        \midrule
        \texttt{offline} & The skill can run without network access or live third-party services. \\
        \texttt{online} & The skill may need public network access or unauthenticated live services. \\
        \texttt{secured} & The skill depends on credentials, API keys, logged-in sessions, or private services. \\
        \bottomrule
    \end{tabularx}
\end{table}

\subsection{Schema of Profiling Artifacts}
\label{app:profiling-artifacts}

\begin{table*}[!t]
    \centering
    \small
    \caption{Quality rubric used during skill profiling. These dimensions decide whether a skill is a stable execution unit before it is used for recommendation.}
    \label{tab:profile-quality-rubric}
    \renewcommand{\arraystretch}{1.3}
    \begin{tabularx}{\textwidth}{>{\raggedright\arraybackslash}p{0.18\textwidth} X X}
        \toprule
        Criterion & Positive Evidence & Negative Evidence \\
        \midrule
        \texttt{consistency} & The skill name, description, instructions, scripts, and referenced resources describe the same capability and compatible assumptions. & The package mixes unrelated goals, contradicts itself, or points to resources that imply a different task. \\
        \texttt{completeness} & The skill provides enough steps, prerequisites, resources, and expected outputs for an agent to execute the capability. & Critical setup, commands, files, inputs, or success conditions are missing. \\
        \texttt{orientation} & The content is written as reusable task-execution guidance for an agent. & The content is mainly background prose, marketing text, a one-off answer, or generic advice without actionable procedure. \\
        \bottomrule
    \end{tabularx}
\end{table*}

\begin{table*}[!t]
    \centering
    \small
    \caption{Verifiability rubric used during skill profiling. A skill passes this profile only when it can support reproducible task construction and objective checking at reasonable cost.}
    \label{tab:profile-verifiability-rubric}
    \renewcommand{\arraystretch}{1.3}
    \begin{tabularx}{\textwidth}{>{\raggedright\arraybackslash}p{0.31\textwidth} X X}
        \toprule
        Criterion & Positive Evidence & Negative Evidence \\
        \midrule
        \texttt{success\_verifiability} & The skill has observable outputs, tests, files, service states, or metrics that can determine success with low ambiguity. & Success depends mainly on subjective preference, hidden state, vague quality judgments, or manual interpretation. \\
        \texttt{environment\_controllability} & The required runtime, tools, services, data, and permissions can be reproduced in a sandbox or benchmark container. & The skill depends on unavailable hardware, unstable external services, private accounts, or uncontrolled live state. \\
        \texttt{task\_constructability} & Concrete task instances, inputs, expected outputs, and verifiers can be created from the skill at reasonable cost. & The skill is too open-ended, too broad, or too expensive to convert into bounded benchmark tasks. \\
        \bottomrule
    \end{tabularx}
\end{table*}

\begin{table*}[!t]
    \centering
    \small
    \caption{Skill categories assigned during profiling. Categories are multi-label and describe the primary capability of a skill.}
    \label{tab:profile-category-enum}
    \renewcommand{\arraystretch}{1.28}
    \begin{tabularx}{\textwidth}{>{\raggedright\arraybackslash}p{0.34\textwidth} X}
        \toprule
        Category & Meaning \\
        \midrule
        \texttt{DataAndAI} & Data processing, analytics, machine learning, model evaluation, notebooks, vector search, and AI application workflows. \\
        \texttt{Development} & Software implementation, refactoring, debugging, build systems, dependency management, and repository-level code work. \\
        \texttt{Testing} & Test writing, repro construction, benchmark execution, CI failure triage, coverage checks, and verifier-oriented workflows. \\
        \texttt{DevOps} & Shell operations, containers, servers, deployment, observability, networking, package managers, and infrastructure configuration. \\
        \texttt{AgentMeta} & Agent skills, prompts, tool orchestration, memory, MCP setup, context engineering, and agent-harness behavior. \\
        \texttt{ProfessionalDomainKnowledge} & Specialized professional workflows such as legal, finance, medicine, business operations, education, or policy analysis. \\
        \texttt{NaturalScienceKnowledge} & Scientific or mathematical workflows in areas such as physics, chemistry, biology, geoscience, and quantitative modeling. \\
        \texttt{Tools} & Concrete use of third-party applications, APIs, CLIs, file formats, document tools, spreadsheet tools, and presentation tools. \\
        \bottomrule
    \end{tabularx}
\end{table*}

\begin{itemize}
    \item \texttt{SkillEnvironmentRubric}: the runtime-requirement profile of a skill. It records supported operating systems through \texttt{os}, the allowed write boundary through \texttt{write\_scope} using the levels in Table~\ref{tab:profile-write-scope}, privilege assumptions through \texttt{privilege}, external dependency level through \texttt{externality} using the levels in Table~\ref{tab:profile-externality}, required environment variables through \texttt{envs}, command-line tools through \texttt{bins}, MCP servers through \texttt{mcps}, and an evidence explanation through \texttt{reason}.
    \item \texttt{SkillQualityRubric}: the quality rubric supplied to the profiling agent. It contains the three rubric dimensions \texttt{consistency}, \texttt{completeness}, and \texttt{orientation}, defined in Table~\ref{tab:profile-quality-rubric}, each paired with a free-text reason.
    \item \texttt{SkillVerifiabilityRubric}: the verifiability rubric supplied to the profiling agent. It contains \texttt{success\_verifiability}, \texttt{environment\_controllability}, and \texttt{task\_constructability}, defined in Table~\ref{tab:profile-verifiability-rubric}, each paired with a free-text reason.
    \item \texttt{SkillEvaluationRubric}: the flattened rubric rendered into the profiling prompt. It combines runtime-requirement fields, quality fields, verifiability fields, category labels from Table~\ref{tab:profile-category-enum}, and category rationale into one schema so that the prompt exposes all decision criteria together.
\end{itemize}

\subsection{Schema of Recommendation Artifacts}

\begin{itemize}
    \item \texttt{skill\_names} (\texttt{list[str]}): the list of skill names recommended to the solver agent. Each name must exactly correspond to a real skill directory under the candidate skill root, and duplicates are not allowed. An empty list is allowed only after effective search confirms that no relevant reusable skill exists.
    \item \texttt{optimized\_context} (\texttt{str}): concise skill-use guidance for the solver agent. It should explain which task stage each selected skill covers, how the skills should be combined, obvious coverage gaps, and usage boundaries. It must not directly complete the task, output the final answer, repeat the search trace, or copy long skill content.
\end{itemize}

\begin{table}[!ht]
    \centering
    \small
    \caption{Judge signal types used by task attribution. Each type identifies the primary evidence source for judging a subtask outcome.}
    \label{tab:judge-enum}
    \renewcommand{\arraystretch}{1.35}
    \begin{tabularx}{\linewidth}{>{\raggedright\arraybackslash}p{0.24\linewidth} @{\hspace{0.10\linewidth}} X}
        \toprule
        Judge Type & Meaning \\
        \midrule
        \texttt{environment} & Primarily judged by observable environment feedback. \\
        \texttt{human} & The result depends on human preference or manual review. \\
        \texttt{unknown} & No clear judgment signal exists. \\
        \bottomrule
    \end{tabularx}
\end{table}
\subsection{Schema of Attribution Artifacts}

\begin{itemize}
    \item \texttt{subtasks} (\texttt{list[Subtask]}): the list of subtasks extracted from this trajectory, containing at least one element.
    \item \texttt{goal} (\texttt{str}): the independent objective of the subtask.
    \item \texttt{summary} (\texttt{str}): the factual summary of the subtask.
    \item \texttt{exploration} (\texttt{str | null}): reusable exploration produced in the subtask; \texttt{null} if there is no reusable content worth retaining.
    \item \texttt{exploration\_reason} (\texttt{str}): the explanation for \texttt{exploration}.
    \item \texttt{judge} (\texttt{enum}): the primary judgment signal type used by the subtask, with values shown in Table~\ref{tab:judge-enum}.
    \item \texttt{judge\_reason} (\texttt{str}): the evidence explanation for selecting this judge type.
    \item \texttt{attribution} (\texttt{enum}): the final result and main-cause category of the subtask, with values shown in Table~\ref{tab:attribution-enum}.
    \item \texttt{attribution\_reason} (\texttt{str}): the evidence explanation for selecting this attribution.
    \item \texttt{skill\_linked} (\texttt{str | null}): the single skill name associated with the subtask.
    \item \texttt{skill\_refs} (\texttt{list[SkillRef]}): the skill text spans actually relied on by the subtask.
    \item \texttt{skill\_refs[].file\_path} (\texttt{str}): the relative path of the referenced file inside the skill directory.
    \item \texttt{skill\_refs[].start\_line} (\texttt{int | null}): the 1-based starting line number of the referenced knowledge span.
    \item \texttt{skill\_refs[].end\_line} (\texttt{int | null}): the 1-based ending line number of the referenced knowledge span.
    \item \texttt{skill\_refs[].capability} (\texttt{str}): the capability, instruction, or knowledge summary expressed by the referenced span.
    \item \texttt{skill\_refs[].used\_for} (\texttt{str}): how the knowledge span was actually used in the current subtask.
    \item \texttt{ground\_truth\_path} (\texttt{str | null}): the oracle directory path attached by the program in offline oracle mode. It is not directly output by the agent.
\end{itemize}

\begin{table*}[!t]
    \centering
    \small
    \caption{Attribution categories produced for each subtask. Successful categories can trigger skill creation or update, while failed and uncertain categories are skipped during skill evolution.}
    \label{tab:attribution-enum}
    \renewcommand{\arraystretch}{1.35}
    \begin{tabularx}{\textwidth}{>{\raggedright\arraybackslash}p{0.28\textwidth} X @{\hspace{0.025\textwidth}}>{\centering\arraybackslash}p{0.11\textwidth}}
        \toprule
        Attribution Type & Meaning & Evolution Type \\
        \midrule
        \path{success_viewed_skill_but_not_used} & The agent viewed a skill, but the skill did not materially shape the successful path. & Create \\
        \path{success_no_skill_seen} & The agent did not view a skill and still completed the subtask through independent exploration. & Create \\
        \path{success_skill_used_with_extra_exploration} & The agent genuinely relied on a skill, performed extra exploration, and succeeded. & Edit or Create \\
        \path{fail_skill_issue} & The main failure cause lies in the skill itself. & Skip \\
        \path{fail_agent_limit} & The main failure cause lies in the agent. & Skip \\
        \path{fail_client_env} & The main failure cause lies in the client-side environment. & Skip \\
        \path{fail_external_env} & The main failure cause lies in external systems or services. & Skip \\
        \path{fail_unknown_env} & The subtask clearly failed because of the environment, but the evidence cannot distinguish the client environment from the external environment. & Skip \\
        \path{uncertain_human_judge_required} & Human judgment is required but currently unavailable. & Skip \\
        \path{uncertain_environment_judge_inconclusive} & Environment signals exist but are insufficient for complete judgment. & Skip \\
        \path{uncertain_no_judge} & No clear judgment signal exists and the goal is not self-evident enough. & Skip \\
        \bottomrule
    \end{tabularx}
\end{table*}

\begin{table*}[!t]
    \centering
    \small
    \caption{Evolution action types and corresponding operations. Each action describes how reusable exploration is evolved, ranging from editing an existing skill to creating a new skill or skip.}
    \label{tab:action-enum}
    \renewcommand{\arraystretch}{1.45}
    \begin{tabularx}{\textwidth}{>{\raggedright\arraybackslash}p{0.23\textwidth} @{\hspace{0.035\textwidth}} X @{\hspace{0.04\textwidth}}>{\centering\arraybackslash}p{0.12\textwidth}}
        \toprule
        Evolution Action Type & Meaning & Skill Operation \\
        \midrule
        \texttt{error\_fix} & Correct clearly wrong or misleading guidance in an old skill. & Edit \\
        \texttt{knowledge\_addition} & Add missing reusable steps, branches, or fallback guidance to an old skill. & Edit \\
        \texttt{prerequisite\_addition} & Add prerequisites, applicability boundaries, warnings, or guardrails. & Edit \\
        \texttt{create\_skill} & Create a new independent skill. & Create \\
        \texttt{skip} & Do not update. & Skip \\
        \bottomrule
    \end{tabularx}
\end{table*}
\subsection{Schema of Evolution Artifacts}

\begin{itemize}
    \item \texttt{request\_dir\_name} (\texttt{str}): the working directory name of the evolution request.
    \item \texttt{target\_skill\_name} (\texttt{str | null}): the old skill name corresponding to an edit request; \texttt{null} for a create request.
    \item \texttt{subtasks} (\texttt{list[Subtask]}): the subtasks supporting this evolution request.
    \item \texttt{actions} (\texttt{list[Action]}): the list of actions returned by the agent.
    \item \texttt{action\_type} (\texttt{enum}): the evolution action category, with values shown in Table~\ref{tab:action-enum}.
    \item \texttt{rationale} (\texttt{str}): the reason for executing the action.
    \item \texttt{summary} (\texttt{str | null}): the edit summary when editing an old skill; \texttt{null} for creation or skip.
    \item \texttt{skill\_dir\_path} (\texttt{str | null}): the absolute directory path of a newly created skill; \texttt{null} when editing an old skill or skipping.
\end{itemize}

\subsection{Aggregation of Subtasks}
Before evolution, the system first merges all subtasks payloads in a batch, and then checks the subtasks one by one. A subtask enters evolution only when it satisfies two conditions:
\begin{itemize}
    \item \texttt{exploration} is nonempty, because we only evolve explorations performed by the agent into skills.
    \item \texttt{attribution} belongs to the three successful exploration categories in Table~\ref{tab:attribution-enum}. Failed and uncertain subtasks do not trigger library updates.
\end{itemize}

After aggregation, successful exploration without an editable linked skill is placed into a create request. For the attribution value \texttt{success\_skill\_used\_with\_extra\_\allowbreak exploration}, successful exploration with a nonempty \texttt{skill\_linked} is grouped by linked skill into multiple edit requests. Each edit request corresponds to exactly one old skill.

\section{Implementation Details}
\label{app:implementation-details}
\subsection{Skill Corpus Validation and Profiling}
\label{app:skill-corpus-validation}
\ours validates, deduplicates, and profiles collected Agent Skills before they enter recommendation or benchmark task synthesis. The profiling process produces three structured views for each accepted skill package: a runtime-requirement profile, a quality profile, and a verifiability profile.

\paragraph{Format validation.}
\ours first checks whether a collected candidate is a valid Agent Skill package. We use Anthropic's official skill validation script\footnote{\url{https://github.com/anthropics/skills/blob/57546260929473d4e0d1c1bb75297be2fdfa1949/skills/skill-creator/scripts/quick_validate.py}} to validate the required \texttt{SKILL.md} format and package structure, and discards malformed candidates before downstream indexing.

\paragraph{Deduplication.}
\ours then removes duplicates among candidates whose \texttt{SKILL.md} files contain exactly the same content. For each duplicate group, the system keeps the copy with the earliest GitHub commit timestamp and treats later copies as duplicates. This preserves the earliest observed source while removing redundant packages from the searchable corpus.

\paragraph{Agentic Skill Profiling.}
For each remaining skill directory, \ours launches Claude Code through the Claude Agent SDK\footnote{\url{https://github.com/anthropics/claude-agent-sdk-python}} with the skill directory as the working scope. We use a Claude Code agent rather than a single-document LLM call because the profiler can inspect the complete skill package, including scripts, references, assets, and auxiliary files, instead of relying only on \texttt{SKILL.md}. This exposes dependencies, applicability boundaries, and missing resources that are often distributed across the directory. The Claude Code process receives only Grep, Glob, Read, and StructuredOutput tools. The final \texttt{StructuredOutput} payload is parsed into the skill profiles described in Appendix~\ref{app:approach-details}.

\subsection{Lifecycle of Harbor Evaluation Framework}
\ours implements benchmark evaluation based on Harbor by integrating Skill Recommendation, Subtask Distillation, and Skill Evolution into the Harbor workflow.

A Harbor job first parses the task collection and expands each task instance into a trial. The main lifecycle of a trial includes creating the trial working directory, starting the task container, running agent setup, passing the instruction to the solver agent, downloading agent logs and sessions, running the verifier, stopping the container, recording the final trial result, and triggering the trial-end hook.

This order determines the following implementation logic of \ours:
\begin{itemize}
    \item The recommendation stage must run before the solver agent starts, because it controls which skills are installed into the agent-visible directory and appends compressed skill-use guidance to the task instruction.
    \item Attribution and skill evolution must run after the trial ends, because they depend on the complete agent session and verifier result. At that point, the task container has already stopped, so both stages run on the host side.
\end{itemize}

\ours implements the above mechanism using the trial lifecycle hook exposed by Harbor. When the trial-end hook is triggered, it can already access the trial result, agent artifacts, and verifier artifacts, which match the inputs required by attribution and evolution. If a failed trial will still be retried by Harbor itself, \ours skips attribution and evolution for that attempt and waits until the final attempt finishes, avoiding duplicate writes of intermediate failures into the experience library.

The task-solving process of the solver agent keeps Harbor's original logic, including execution and verifying. \ours only changes the preparation stage before task execution and the attribution stage after task execution, without modifying the solver agent's task-solving workflow.

\subsection{Dataset Preparation and Environment Setup}
When using Harbor's official dataset images, each trial dynamically installs the agent CLI during setup by default. This creates substantial network I/O in concurrent runs and limits the throughput of large experiment batches. To improve experimental efficiency, \ours therefore prebuilds experiment images with the agent CLI on top of the original task images and skips repeated installation at runtime.

The prebuild process first downloads the Harbor dataset, reads the image definition for each task, builds a new image with a preinstalled agent from the original task image as the base image, and uses the built image in the task configuration. The prebuilt image fixes \texttt{nvm}~\texttt{0.40.4}, Node.js~\texttt{22}, and Codex CLI~\texttt{0.125.0}. This approach reduces repeated downloads in concurrent experiments and shortens the trial preparation time.

\subsection{Experiment Configuration and Orchestration}
\ours provides a lightweight launcher outside Harbor. The launcher uses YAML to manage both native Harbor configuration and \ours-specific extended configuration, registers different modules, and makes experimental setup convenient.

This design allows baselines, recommendation, online evolution, and offline evolution to share the same launcher. Different experiments only need to change YAML, without modifying Harbor code or copying multiple execution scripts.

\subsection{Integration of Solver Agent}
We implement our solver agent based on the Codex integration provided by Harbor, but do not modify its task execution logic. We only modify the preparation work before trial execution. Because the image already preinstalls Codex~\texttt{0.125.0}, agent setup no longer installs the CLI and only creates the necessary agent directories. During formal execution, Codex still runs the task instruction in execution mode, with JSON logging and the unified terminal execution tool enabled so that Harbor can download the session and execution status.

Codex's built-in system skills and plugins may inject extra prompt content, interfering with the measurement of the \ours skill library. The system first initializes the agent home so that Codex discovers system skills and plugins; it then generates the agent configuration according to the experiment setting and marks system skills and plugins as disabled.

\subsection{Integration of Skill Recommendation}
The goal of the recommendation stage is to select a small set of skills that are most relevant to the task and least redundant before the solver agent executes. The agent itself does not solve the task; it only searches the candidate skill library, reads candidate skill documents and resources, and generates skill usage guidance for the solver agent.

\begin{enumerate}
    \item The candidate skill library is mounted into the task container through Harbor as a read-only directory. The agent's \texttt{cwd} is set to the candidate skill root, not the benchmark task workspace.
    \item An isolated recommendation environment is created. The recommendation stage uses a temporary agent home and a temporary output directory. System skills and plugins are disabled in the same controlled way as for the solver, and the candidate skill root is used only as the trusted directory for recommendation.
    \item The recommendation prompt is rendered and executed. The recommendation stage reuses the solver agent's model and CLI parameters and runs with bypass permissions.
    \item The recommendation stage writes the JSON schema, structured output file, command log, and recommendation session. The structured output is parsed and validated; if the output is missing or malformed, it is retried up to three times. Logs, intermediate outputs, final outputs, and the recommendation session are downloaded to the host trial directory.
    \item If recommendation succeeds and returns a nonempty skill-name list, the system copies only those skill directories into the solver agent's skill directory, and the concise usage guidance generated by the agent is appended to the end of the task instruction as the solver agent's skill usage context. If the recommendation stage repeatedly fails to produce valid output, or if installing the selected skills fails, the system records the error and falls back to copying all candidate skills, allowing the benchmark trial to continue.
    \item After recommendation ends, the temporary recommendation directory, temporary agent home, and temporary credential files inside the container are cleaned up.
\end{enumerate}

\subsection{Integration of Task Attribution}
This stage turns a complete agent trajectory and verifier result into structured subtasks. It does not simply compress the original session into text; instead, it resumes the original Codex session so that the agent continues reasoning inside the native agent harness context. This prevents loss of contextual details, and because commercial models often encrypt chain-of-thought, the resume mechanism is the only way to avoid discarding that context.

\begin{enumerate}
    \item When the trial-end hook is triggered, Harbor has already downloaded agent artifacts, run the verifier, and stopped the task container. The host-side session and verifier artifacts can therefore be obtained.
    \item An isolated working path and a new agent home are created for each trial. The original solver agent session, visible skills, and related artifacts are copied.
    \item Because each trial is expected to produce exactly one agent session file, the system reads the session id needed for resume from that session file and runs Codex resume with the isolated working path and agent home to restore the context at that time. The prompt is appended as a new user prompt through standard input; resume does not replace the original system prompt, preserving native session context management.
    \item Verifier evidence is provided in a controlled way, mainly in two modes:
    \begin{itemize}
        \item Online mode provides only task-level test counts, including the total, passed, and failed counts. These counts can come from CTRF reports, pytest summaries, benchmark-specific JSON, or Harbor reward.
        \item Offline oracle mode can additionally expose paths to the solution, verifier tests, and verifier stdout, but the prompt still forbids writing gold answers, private constants, canary strings, one-off paths, or exact ground-truth outputs into reusable exploration.
    \end{itemize}
    \item The output is validated and the resumed session is archived. The attribution stage uses a structured schema and writes the last message to a JSON output file. Missing output, malformed output, runtime timeout, or missing resumed session artifacts are treated as retryable output errors and are retried up to three times. After validation passes, artifacts are retained.
\end{enumerate}

\subsection{Integration of Skill Evolution}
The evolution stage runs after attribution stage, and it only consumes structured subtasks. The system uses a unified local agent home to store credentials and agent sessions, while each evolution request has an independent working directory to isolate the read/write scope.
\label{app:case-apache-transfer}
\begin{figure*}[!t]
    \centering
    \includegraphics[width=\textwidth]{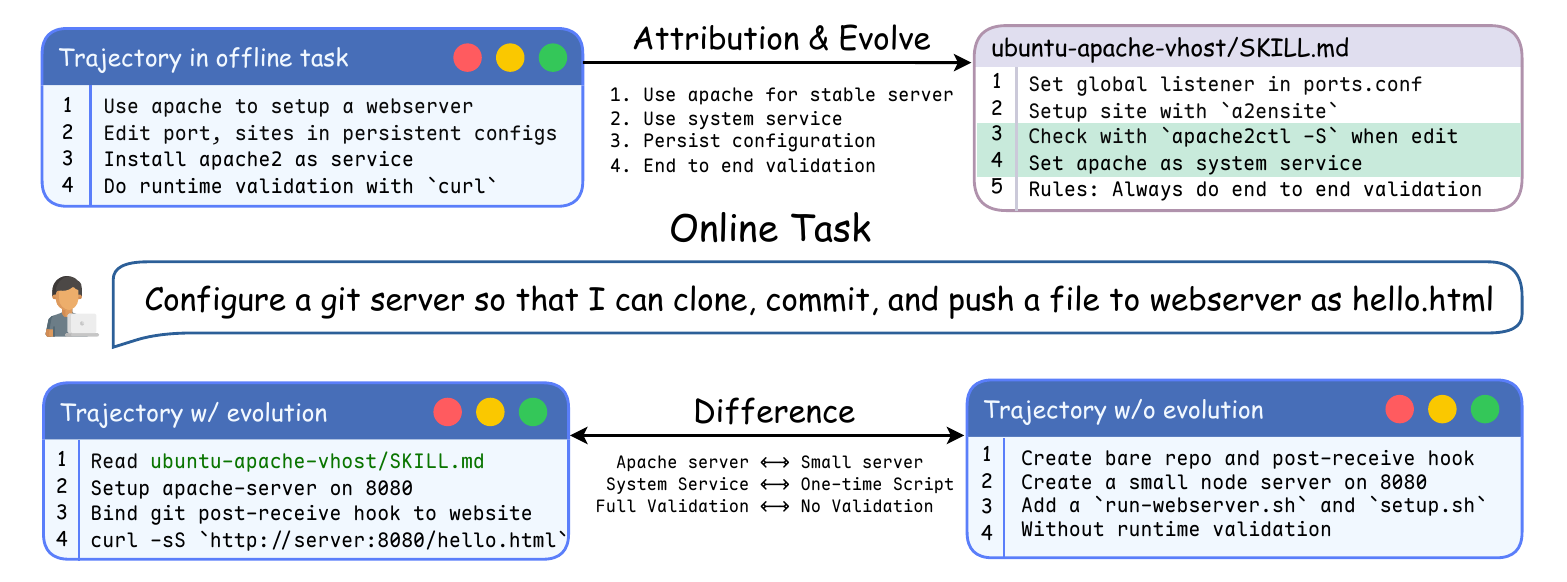}
    \caption{Representative offline-transfer case. A skill evolved from an Apache website task transfers persistent-service setup and end-to-end validation to an unseen Git-server deployment task.}
    \label{fig:offline-transfer-case}
\end{figure*}
\begin{enumerate}
    \item Subtasks are aggregated and evolution requests are constructed. Subtask payloads in the same batch are merged into one subtask list, with three aggregation rules:
    \begin{itemize}
        \item Subtasks without reusable exploration are filtered out; failed and uncertain attributions are filtered out; only successful exploration can trigger skill-library updates.
        \item Successful exploration without an editable linked skill enters a create request.
        \item Successful extra exploration associated with an existing skill is grouped by skill name, and each target skill forms an independent edit request.
    \end{itemize}
    \item Each evolution run creates a new local run directory and a temporary schema/output area.
    \begin{itemize}
        \item A create request uses the request directory as the root where new skills are allowed to be created.
        \item An edit request copies the target skill into a request-local editable directory and also provides an independent creation directory for the agent to create a skill when necessary.
        \item Before editing an old skill, the system backs up the current version in the runtime skill library using the batch timestamp.
    \end{itemize}
    \item Create requests use the create system prompt and create user prompt, and edit requests use the edit system prompt and edit user prompt; Appendix~\ref{app:prompt-skill-evolution} lists all four prompts.
    \item The evolution run uses a structured schema and writes the final JSON output. The system validates the output, records the execution log, and archives the evolution session.
\end{enumerate}

\section{Case Study}
\label{app:case-study}
\newcommand{\svCaseMarkdownHeadingTwo}[1]{%
  \par\smallskip{\sffamily\bfseries\color{svPromptBlue}#1}\par
}

\newcommand{\svCaseMarkdownHeadingThree}[1]{%
  \par\smallskip{\sffamily\bfseries #1}\par
}

\newcommand{\svCaseMarkdownHeadingFour}[1]{%
  \par\smallskip{\sffamily\bfseries\itshape #1}\par
}

\newlength{\svCaseCodeSpanMeasure}
\newif\ifsvCaseCodeSpanHasSpace
\makeatletter
\def\svCaseCodeSpanCheckSpace#1{%
  \svCaseCodeSpanHasSpacefalse
  \svCaseCodeSpanCheckSpaceAux#1 \svCaseCodeSpanSentinel
}
\def\svCaseCodeSpanCheckSpaceAux#1 #2\svCaseCodeSpanSentinel{%
  \def\svCaseCodeSpanRest{#2}%
  \ifx\svCaseCodeSpanRest\@empty
  \else
    \svCaseCodeSpanHasSpacetrue
  \fi
}
\makeatother

\newcommand{\svCaseMarkdownCodeSpan}[1]{%
  \textnormal{{\ttfamily\color{svPromptAccentColor}\textasciigrave{}%
  \begingroup
    \def\markdownRendererUnderscore{\_\hspace{0pt}}%
    \def\markdownRendererPercentSign{\%\hspace{0pt}}%
    \def\markdownRendererLeftBrace{\{\hspace{0pt}}%
    \def\markdownRendererRightBrace{\}\hspace{0pt}}%
    \settowidth{\svCaseCodeSpanMeasure}{{\ttfamily #1}}%
    \svCaseCodeSpanCheckSpace{#1}%
    \ifsvCaseCodeSpanHasSpace
      #1%
    \else\ifdim\svCaseCodeSpanMeasure>.35\linewidth
      \seqsplit{#1}%
    \else
      #1%
    \fi\fi
  \endgroup
  \textasciigrave{}}}%
}

\newcommand{\svCaseMarkdownThematicBreak}{%
  \par\medskip\noindent{\color{svPromptBlue!45}\leaders\hbox{\rule[0.45ex]{5pt}{0.35pt}\hspace{3pt}}\hfill\kern0pt}\par\medskip
}

\newcommand{\svCaseMarkdownSoftLineBreak}{%
  \par
}

\newcommand{\svCaseMarkdownUlBegin}{%
  \begin{itemize}[leftmargin=1.05em,labelsep=0.35em]%
}

\newcommand{\svCaseMarkdownUlBeginTight}{%
  \begin{itemize}[leftmargin=1.05em,labelsep=0.35em,noitemsep]%
}

\newcommand{\svCaseMarkdownOlBegin}{%
  \begin{enumerate}[leftmargin=1.45em,labelsep=0.35em]%
}

\newcommand{\svCaseMarkdownOlBeginTight}{%
  \begin{enumerate}[leftmargin=1.45em,labelsep=0.35em,noitemsep]%
}

\makeatletter
\newcommand{\svCaseMarkdownUlItem}{%
  \@ifnextchar\markdownRendererStrongEmphasis
    {\svCaseMarkdownUlItemStrong}
    {\item}%
}

\def\svCaseMarkdownUlItemStrong\markdownRendererStrongEmphasis#1:{%
  \edef\svCaseMarkdownField{\detokenize{#1}}%
  \edef\svCaseMarkdownSkipField{\detokenize{skill\markdownRendererUnderscore{}refs}}%
  \ifnum\pdfstrcmp{\svCaseMarkdownField}{\svCaseMarkdownSkipField}=0
    \expandafter\svCaseMarkdownGobbleUlItem
  \else
    \item\markdownRendererStrongEmphasis{#1}:%
  \fi
}

\long\def\svCaseMarkdownGobbleUlItem#1\markdownRendererUlItemEnd{}

\newcommand{\svCaseMarkdownUlItemEnd}{}
\makeatother

\newcommand{\svcasemarkdownfile}[2]{%
  \begin{tcolorbox}[
    enhanced standard jigsaw,
    breakable,
    width=\linewidth,
    colback=svPromptBlueLight,
    colframe=svPromptBlue,
    colbacktitle=svPromptBlue,
    coltitle=white,
    arc=4pt,
    fonttitle=\bfseries\sffamily,
    fontupper=\small,
    title={#1},
    before upper={%
      \setminted{%
        breaklines,
        breakanywhere,
        breakautoindent=false,
        breakindent=0pt,
        breaksymbolleft={},
        breaksymbolright={},
        tabsize=2%
      }%
    }
  ]
  \markdownInput[
    html=false,
    fencedCode=true,
    rendererPrototypes={
      headingTwo={\svCaseMarkdownHeadingTwo{##1}},
      headingThree={\svCaseMarkdownHeadingThree{##1}},
      headingFour={\svCaseMarkdownHeadingFour{##1}},
      codeSpan={\svCaseMarkdownCodeSpan{##1}},
      softLineBreak={\svCaseMarkdownSoftLineBreak},
      ulBegin={\svCaseMarkdownUlBegin},
      ulBeginTight={\svCaseMarkdownUlBeginTight},
      ulItem={\svCaseMarkdownUlItem},
      ulItemEnd={\svCaseMarkdownUlItemEnd},
      ulEnd={\end{itemize}},
      ulEndTight={\end{itemize}},
      olBegin={\svCaseMarkdownOlBegin},
      olBeginTight={\svCaseMarkdownOlBeginTight},
      olEnd={\end{enumerate}},
      olEndTight={\end{enumerate}},
      thematicBreak={\svCaseMarkdownThematicBreak}
    }
  ]{#2}%
  \end{tcolorbox}%
}

\subsection{Offline Skill Transfer from TB-Pro to Terminal-Bench 2.0}

As shown in Figure~\ref{fig:offline-transfer-case}, this group of cases provides a representative example of what is transferred by offline evolution. The source trajectories implement Apache-backed websites and contribute reusable knowledge about persistent Apache configuration, service installation, and end-to-end runtime validation. These lessons are distilled into the ubuntu-apache-vhost skill rather than stored as raw trajectories. 

On the unseen \texttt{configure-git-webserver} task, the evolved run does not copy the source solution. Instead, it reuses the operational pattern: deploy the web endpoint with a stable Apache service, connect the Git post-receive hook to the served directory, and validate the full path by requesting the final URL. The baseline run builds a bare repository and a lightweight Node server, but it lacks persistent service setup and final runtime validation. The case illustrates the type of transfer that \ours is designed to preserve: not task-specific constants or answers, but reusable execution invariants that improve reliability on a different task.

This case shows the creation of the \texttt{ubuntu-apache-vhost} skill when running offline evolution on the task of Terminal-Bench Pro. During task execution, the agent persists configuration parameters through Apache configuration files, configures the web root, and validates both the configuration and site availability. These exploratory steps are extracted into subtasks, and reusable knowledge such as service startup and runtime validation is converted into a skill.
\svcasemarkdownfile{Creation of skill from \texttt{\seqsplit{configure-apache-logging-and-rate-limit} }}{assets/case_study/apache-logging-ratelimit.md}

This case shows the editing of the \texttt{ubuntu-apache-vhost} skill when running offline evolution on the task of Terminal-Bench Pro. During task execution, the agent follows the skill to launch an Apache service, discovers that \texttt{systemctl} is unavailable, and falls back to the \texttt{service} command. These explorations are extracted into subtasks, and the evolution step updates the skill with a fallback strategy for system service installation while strengthening its runtime validation procedure.
\svcasemarkdownfile{Editing of skill from \texttt{\seqsplit{configure-apache-analytics-virtualhost}}}{assets/case_study/apache-analytics.md}

This case shows the skill transfers to unseen task of Terminal-Bench 2.0, with recommendation enabled. The \texttt{ubuntu-apache-vhost} skill is recommended for persistent HTTP serving and runtime validation. During problem solving, the agent reuses this operational pattern by launching the HTTP endpoint through Apache, installing it as a system service via the \texttt{service} command, and repeatedly validating the service both during intermediate steps and at the end of the task. This demonstrates that the skill captures reusable operational knowledge rather than a task-specific answer, and that such knowledge can transfer successfully to an unseen task.
\svcasemarkdownfile{Transfer of skill to \texttt{\seqsplit{configure-git-webserver}}}{assets/case_study/apache-transfer-evolution.md}

This case shows the baseline without skills creates a bare repository, a \texttt{post-receive} deployment hook, and a lightweight Node server, but it does not ensure service stability or end-to-end validation.
\svcasemarkdownfile{w/o skills run on \texttt{\seqsplit{configure-git-webserver}}}{assets/case_study/apache-transfer-baseline.md}

\subsection{Online Evolution on SWE-Bench Pro}

This case shows the online evolution run on the NodeBB task of SWE-Bench Pro, in which multiple recommended skills support different parts of the repair. Route wiring, authenticated reproduction, Redis bootstrap, and API error serialization are separated into distinct attributed subtasks, and only the reusable successful parts are eligible for editing skills.

\svcasemarkdownfile{Online evolution run on \texttt{\seqsplit{nodebb-invite-api}}}{assets/case_study/nodebb-group-invite.md}

\clearpage
\section{Prompts}
\label{app:prompts}
% This file renders prompt Markdown files from assets/prompts with tcolorbox's
% minted listing engine. Prompt blocks are non-floating, single-column, and
% breakable in the two-column ACL appendix.

\raggedbottom

\newcommand{\svpromptfile}[3][]{%
  \begingroup
  \renewcommand{\theFancyVerbLine}{\scriptsize\color{gray!70!black}\arabic{FancyVerbLine}}%
  \tcbinputlisting{%
    enhanced standard jigsaw,
    listing engine=minted,
    minted language=markdown,
    listing file={#3},
    listing only,
    breakable,
    notitle after break,
    lines before break=1,
    width=\linewidth,
    colback=svPromptBlueLight,
    colframe=svPromptBlue,
    colbacktitle=svPromptBlue,
    coltitle=white,
    arc=4pt,
    fonttitle=\bfseries\sffamily,
    title={#2},
    minted options={%
      linenos,
      numbersep=4pt,
      xleftmargin=0.2em,
      breaklines,
      breakanywhere,
      breakautoindent=false,
      breakindent=0pt,
      breaksymbolleft={},
      breaksymbolright={},
      tabsize=2
    },
    #1%
  }%
  \endgroup
}

\newcommand{\svpromptfilewithtextjsonplaceholder}[3][]{%
  \begingroup
  \renewcommand{\theFancyVerbLine}{\scriptsize\color{gray!70!black}\arabic{FancyVerbLine}}%
  \begin{tcolorbox}[
    enhanced standard jigsaw,
    breakable,
    notitle after break,
    lines before break=1,
    width=\linewidth,
    colback=svPromptBlueLight,
    colframe=svPromptBlue,
    colbacktitle=svPromptBlue,
    coltitle=white,
    arc=4pt,
    fonttitle=\bfseries\sffamily,
    title={#2},
    #1%
  ]
  \inputminted[
    linenos,
    numbersep=4pt,
    xleftmargin=0.2em,
    breaklines,
    breakanywhere,
    breakautoindent=false,
    breakindent=0pt,
    breaksymbolleft={},
    breaksymbolright={},
    tabsize=2,
    rangestopbeforestring={```json}
  ]{markdown}{#3}
  \inputminted[
    linenos,
    firstnumber=last,
    numbersep=4pt,
    xleftmargin=0.2em,
    breaklines,
    breakanywhere,
    breakautoindent=false,
    breakindent=0pt,
    breaksymbolleft={},
    breaksymbolright={},
    tabsize=2,
    rangestartstring={```json}
  ]{text}{#3}
  \end{tcolorbox}%
  \endgroup
}

\subsection{Skill Profiling}
\label{app:prompt-skill-profiling}

The profiling stage evaluates one skill directory at a time. The system prompt
requests inspection of the skill's instructions and supporting files, with a
structured report covering runtime requirements, quality signals, and
suitability for automatically verifiable tasks. A skill root is the directory
that contains the target \texttt{SKILL.md} and any scripts, references, or
assets that belong to that skill. Runtime placeholders are kept literal in the
templates below.

% Let the first breakable prompt use the remaining left column before splitting.
% \enlargethispage*{12cm}
\svpromptfile{System Prompt}{assets/prompts/profiling-system.md}

The user prompt supplies the concrete skill root to evaluate. That directory
defines the evidence boundary for the structured evaluation.

\svpromptfile{User Prompt}{assets/prompts/profiling-user.md}

\subsection{Skill Recommendation}
\label{app:prompt-skill-recommendation}

The recommendation stage is used before task execution to select a small set of
skills and produce compact guidance for the downstream solver.

The directory organization is the default form in the benchmark experiments:
each candidate skill is an installable folder containing \texttt{SKILL.md} and
optional supporting files. This form is also used for skills produced by the
skill-creator setting and by skill evolution.

\svpromptfile{System Prompt (Directory Skills)}{assets/prompts/recommendation-system.md}

The markdown-library organization is used only when recommending over the
\ours-curated skill library. In this setting, each candidate is a markdown
file placed under a category directory, and referenced scripts or assets are
not part of the candidate evidence.

\svpromptfile{System Prompt (Markdown Library)}{assets/prompts/recommendation-library-system.md}

Both recommendation variants use the same user prompt. It supplies the skill
library root and the task instruction that should condition skill selection.

\svpromptfile{User Prompt}{assets/prompts/recommendation-user.md}

\subsection{Skill Routing}
\label{app:prompt-skill-routing}

For benchmarking on the SkillRouter dataset, \ours ranks relevant skill
documents for a query. This setting differs from the default
recommendation stage because the output is a ranked document list rather than
solver-facing usage guidance.

\svpromptfile{System Prompt}{assets/prompts/skillrouter-routing-system.md}

\svpromptfile{User Prompt}{assets/prompts/skillrouter-routing-user.md}

\subsection{Post-Execution Attribution}
\label{app:prompt-post-execution-attribution}

After task execution, attribution summarizes the trajectory into subtasks and
assigns outcome responsibility using the execution trace, accessible skills, and
task-level verifier signal. The main prompt is used in both online and offline
evolution settings.

\svpromptfile{User Prompt}{assets/prompts/attribution-user.md}

Offline evolution can additionally expose reference material after the task has
finished. The optional context below is inserted only in that setting. It helps
the attribution step interpret verifier behavior and check whether a successful
exploration is actually correct, while forbidding task answers, private values,
or one-off outputs from becoming reusable skill knowledge.

\svpromptfile{Reference Context Prompt (Offline Evolution Optional)}{assets/prompts/ground-truth.md}

\subsection{Skill Evolution}
\label{app:prompt-skill-evolution}

Skill evolution consumes successful attributed subtasks rather than the full
trajectory. Each evolution request is routed into one of two branches depending
on whether the reusable evidence is associated with an existing skill boundary.

The edit branch is used when a target skill exists. Its system prompt defines
how the pipeline chooses among local edits, prerequisite guards, creation of a
separate skill, and skipping weak evidence.

\svpromptfile{System Prompt for Skill Editing}{assets/prompts/evolution-edit-system.md}

The edit branch receives a user prompt containing the target skill copy, the
directory where new skills may be created if needed, and the JSON subtasks that
support the request.

\svpromptfilewithtextjsonplaceholder{User Prompt for Skill Editing}{assets/prompts/evolution-edit-user.md}

The create branch is used when the attributed exploration is not routed to an
existing skill. Its system prompt defines the decision rule for whether the
evidence is strong enough to become one or more independent skills, or whether
it should be skipped.

\svpromptfile{System Prompt for Skill Creation}{assets/prompts/evolution-create-system.md}

The create branch receives a user prompt containing the creation directory and
the successful subtasks being considered for new skill construction.

\svpromptfilewithtextjsonplaceholder{User Prompt for Skill Creation}{assets/prompts/evolution-create-user.md}

The skill-creator baseline follows a direct trajectory-to-skill setting. Rather
than deriving separate edit or create requests from attributed subtasks, it
resumes the completed trajectory in an editable skill workspace and keeps only
reusable, transferable, and actionable knowledge.

\svpromptfile{User Prompt for skill-creator Baseline}{assets/prompts/evolution-skill-creator-user.md}

\end{document}